\documentclass[10pt,twocolumn,letterpaper]{article}

\usepackage[pagenumbers]{cvpr} 

\usepackage{graphicx}
\usepackage{multirow}
\usepackage{amsmath,amssymb,amsfonts}
\usepackage{booktabs}
\usepackage{siunitx}
\usepackage{adjustbox}
\usepackage[normalem]{ulem}
\usepackage{subcaption}
\usepackage{listings}
\usepackage{pifont}
\usepackage{url}
\usepackage{threeparttable}

\definecolor{cvprblue}{rgb}{0.21,0.49,0.74}
\usepackage[pagebackref,breaklinks,colorlinks,allcolors=cvprblue]{hyperref}

\def\paperID{*****} 
\def\confName{CVPR}
\def\confYear{2026}

\title{SIMMER: \\Cross-Modal Food Image--Recipe Retrieval via MLLM-Based Embedding}

\author{
  \begin{tabular}{c c}
    Keisuke Gomi & Keiji Yanai \\
    {\tt\small gomi-k@mm.inf.uec.ac.jp} & {\tt\small yanai@cs.uec.ac.jp} \\
   [10pt]
    \multicolumn{2}{c}{\normalsize The University of Electro-Communications} \\
    \multicolumn{2}{c}{\normalsize 1-5-1 Chofugaoka, Chofu-shi, Tokyo 182-8585, Japan}
  \end{tabular}
}

\begin{document}

\twocolumn[{
  \renewcommand\twocolumn[1][]{#1}
  \maketitle
  
  \begin{center}
    \captionsetup{type=figure}

    \begin{subfigure}{0.48\linewidth}
        \centering
        \includegraphics[width=\linewidth]{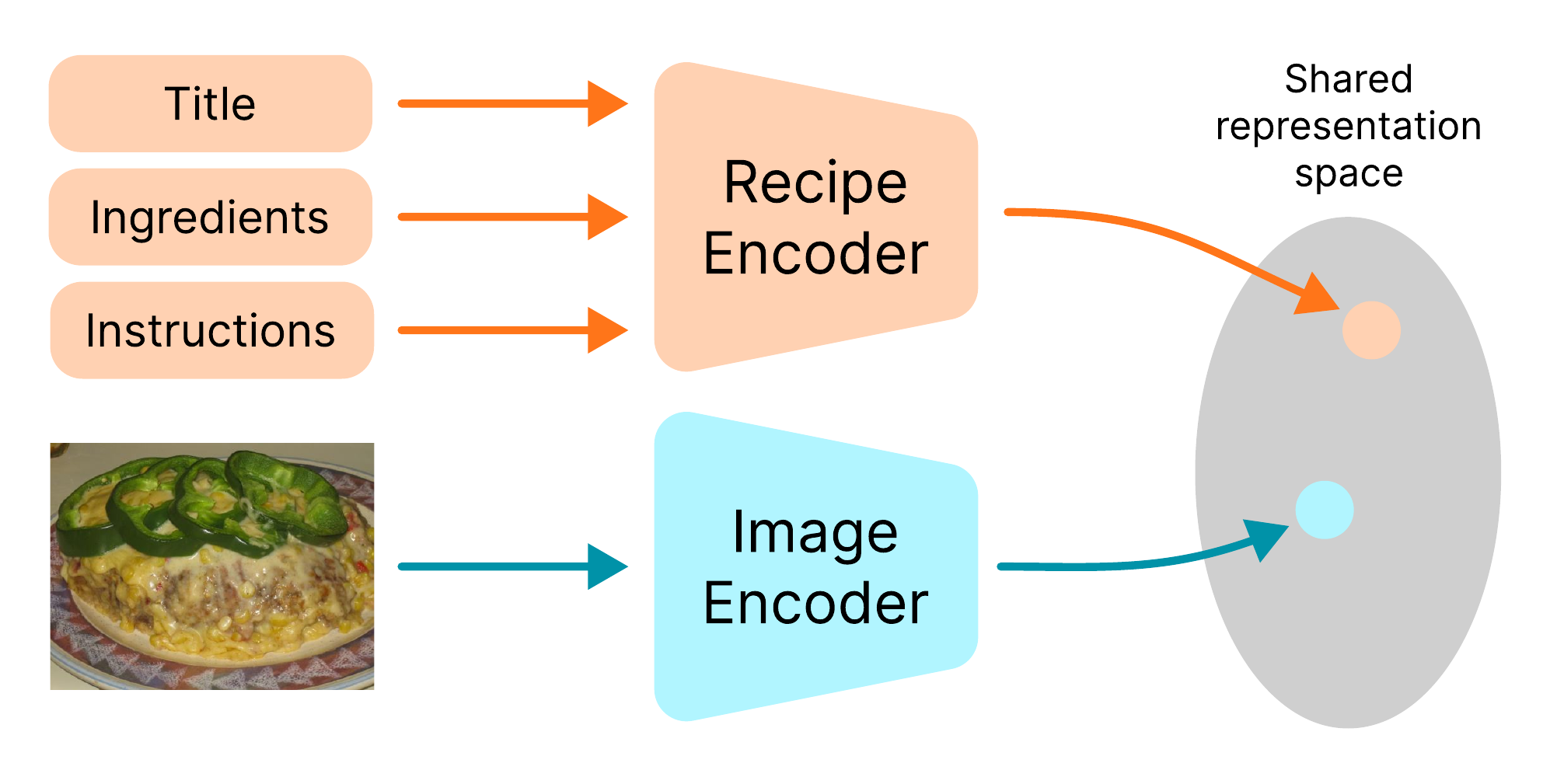}
        \caption{Existing methods}
        \label{fig:overview-existing}
    \end{subfigure}
    \hfill
    \begin{subfigure}{0.48\linewidth}
        \centering
        \includegraphics[width=\linewidth]{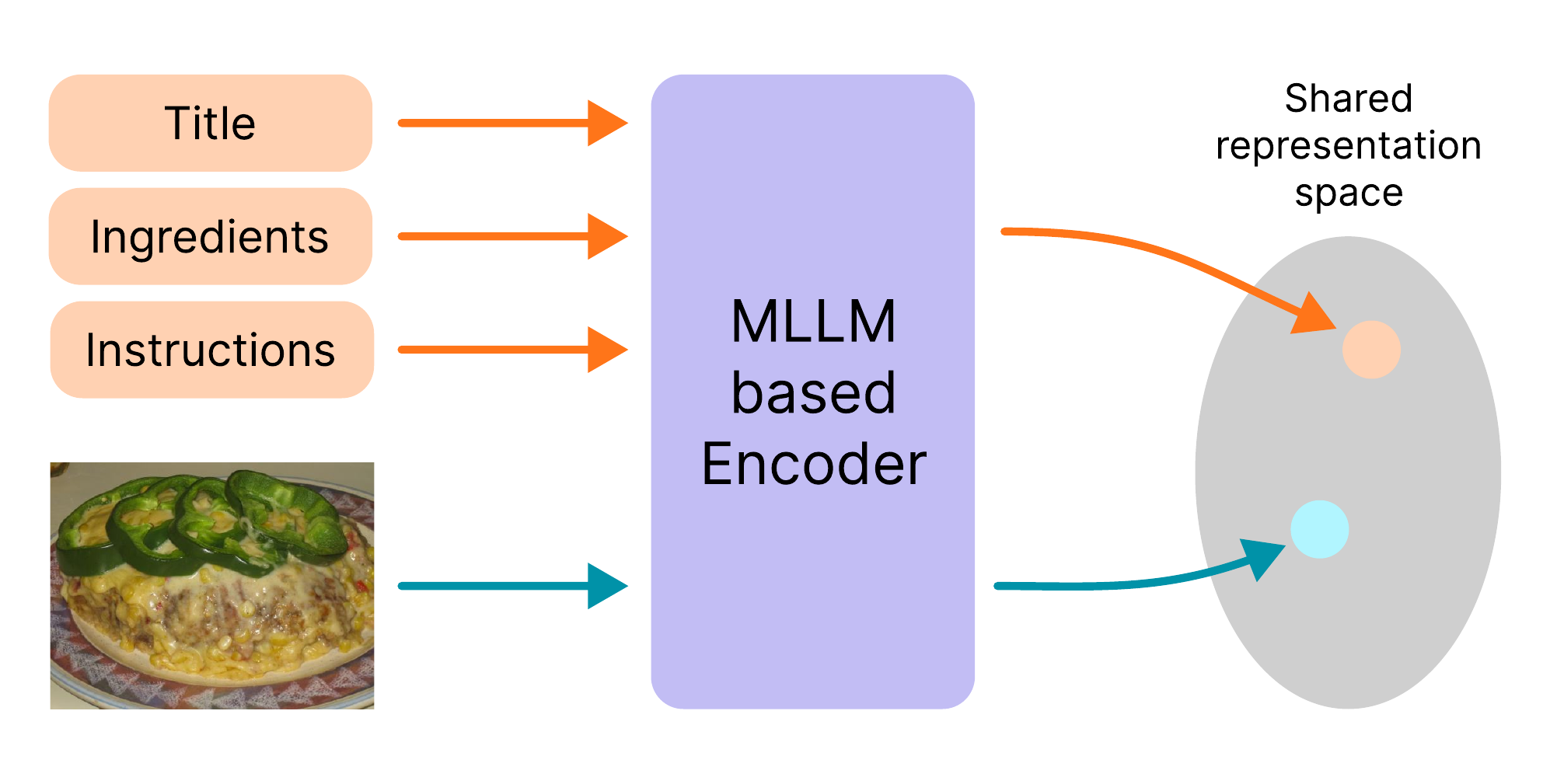}
        \caption{SIMMER (proposed)}
        \label{fig:overview-proposed}
    \end{subfigure}
    
    \caption{Overview of the existing methods and SIMMER}
    \label{fig:overview}
  \end{center}
  \vspace{2em}
}]

\begin{abstract}
Cross-modal retrieval between food images and recipe texts is an important task with applications in nutritional management, dietary logging, and cooking assistance. Existing methods predominantly rely on dual-encoder architectures with separate image and text encoders, requiring complex alignment strategies and task-specific network designs to bridge the semantic gap between modalities. In this work, we propose SIMMER (Single Integrated Multimodal Model for Embedding Recipes), which applies Multimodal Large Language Model (MLLM)-based embedding models, specifically VLM2Vec, to this task, replacing the conventional dual-encoder paradigm with a single unified encoder that processes both food images and recipe texts. We design prompt templates tailored to the structured nature of recipes, which consist of a title, ingredients, and cooking instructions, enabling effective embedding generation by the MLLM. We further introduce a component-aware data augmentation strategy that trains the model on both complete and partial recipes, improving robustness to incomplete inputs. Experiments on the Recipe1M dataset demonstrate that SIMMER achieves state-of-the-art performance across both the 1k and 10k evaluation settings, substantially outperforming all prior methods. In particular, our best model improves the 1k image-to-recipe R@1 from 81.8\% to 87.5\% and the 10k image-to-recipe R@1 from 56.5\% to 65.5\% compared to the previous best method.
\end{abstract}

\section{Introduction}

Food is an essential part of daily life, and interest in healthy eating habits continues to grow worldwide.
In recent years, recipe-sharing platforms and social media have made vast quantities of food images and recipe texts publicly available, driving an increasing demand for techniques that can effectively leverage such multimodal data.
In this context, cross-modal retrieval between food images and recipe texts has emerged as an important research problem.
This task aims to enable bidirectional retrieval: given a food image, the system retrieves the corresponding cooking recipe, and conversely, given a recipe text, it retrieves the matching food image.
Such capabilities have broad practical applications, including nutritional management, dietary logging, and cooking assistance.

Meanwhile, in the fields of computer vision and natural language processing, Multimodal Large Language Models (MLLMs) such as LLaVA~\cite{LLaVA-NeXT-Interleave_Li_LLaVANeXTInterleaveTackling2024} and Qwen2-VL~\cite{Qwen2-VL_Wang_Qwen2VLEnhancing2024} have advanced rapidly, demonstrating remarkable abilities in the integrated understanding of images and text.
More recently, a line of research has explored repurposing MLLMs as embedding models.
Methods such as VLM2Vec~\cite{VLM2Vec_Jiang_VLM2VecTraining2025} and MM-Embed~\cite{MM-EMBED_Lin_MMEmbedUniversal2025} have shown that the rich multimodal knowledge acquired during large-scale pre-training can be effectively transferred to produce high-quality multimodal embeddings.

However, existing methods for cross-modal food image--recipe retrieval predominantly rely on dual-encoder architectures, in which separate image and text encoders are independently designed and trained~\cite{H-T_Salvador_RevampingCrossModal2021,T-Food_Shukor_TransformerDecoders2022,TNLBT_Yang_TransformerBasedCrossModal2023}, as illustrated in Figure~\ref{fig:overview-existing}.
To our knowledge, MLLMs have not yet been applied to this task.
Moreover, dual-encoder approaches typically require elaborate alignment strategies and task-specific network designs to bridge the semantic gap between the visual and textual modalities.
For instance, ACME~\cite{ACME_Wang_LearningCrossModal2019} and R$^2$GAN~\cite{R2GAN_Zhu_R2GANCrossModal2019} employ adversarial training to align the feature distributions across modalities.
Meanwhile, H-T~\cite{H-T_Salvador_RevampingCrossModal2021} and T-Food~\cite{T-Food_Shukor_TransformerDecoders2022} introduce hierarchical Transformer architectures specifically tailored to the structured nature of recipe texts comprising titles, ingredients, and instructions.

In this work, we propose SIMMER (Single Integrated Multimodal Model for Embedding Recipes), which applies MLLM-based embedding models---specifically VLM2Vec-V1~\cite{VLM2Vec_Jiang_VLM2VecTraining2025} and VLM2Vec-V2~\cite{VLM2Vec-V2_Meng_VLM2VecV2Advancing2025}---to cross-modal food image--recipe retrieval, as depicted in Figure~\ref{fig:overview-proposed}.
Since MLLMs are pre-trained on large-scale multimodal corpora, they inherently possess the ability to process both images and text within a unified framework.
By leveraging this pre-trained knowledge, SIMMER eliminates the need for the complex alignment mechanisms and specialized architectures that have been essential in prior work, while still obtaining effective cross-modal embeddings.
We validate the effectiveness of SIMMER through extensive experiments on Recipe1M~\cite{Recipe1M_Salvador_LearningCrossModal2017}, a large-scale benchmark dataset for food image--recipe retrieval.

The main contributions of this paper are summarized as follows:
\begin{itemize}
    \item We are the first to apply MLLM-based embedding models to cross-modal food image--recipe retrieval, demonstrating that a single unified encoder can replace the conventional dual-encoder paradigm.
    \item We propose a method for encoding structured recipe data---consisting of a title, ingredients, and cooking instructions---into a format suitable for MLLM-based embedding generation, along with a component-aware data augmentation strategy to improve robustness to incomplete recipes.
    \item SIMMER achieves state-of-the-art performance on the Recipe1M dataset, substantially outperforming all existing methods across both image-to-recipe and recipe-to-image retrieval tasks.
\end{itemize}

\section{Related work}
\subsection{Cross-Modal Image--Text Embeddings}

Cross-modal embedding of images and text has long been an active area of research.
Early work on visual--semantic embeddings pioneered the idea of projecting images and text into a shared vector space.
DeViSE~\cite{DeViSE_Frome_DeViSEDeep2013} trained a linear mapping from CNN image features to a word-embedding space learned by a skip-gram language model, enabling zero-shot recognition of unseen object categories.
Kiros et al.~\cite{VSE_Kiros_UnifyingVisualSemantic2014} extended this idea by jointly learning image and sentence encoders with a contrastive ranking objective, establishing the Visual Semantic Embedding (VSE) framework widely adopted in subsequent research.
VSE++~\cite{VSEpp_Faghri_VSEImproved2018} improved training efficiency by introducing hard-negative mining into the ranking loss.
SCAN~\cite{SCAN_Lee_StackedCross2018} further advanced fine-grained matching by employing stacked cross-attention between image regions detected by object detectors and individual words, achieving strong performance on image--text retrieval benchmarks.
Beyond dual-encoder designs, a line of work explored early-fusion Transformer architectures for vision--language understanding.
ViLBERT~\cite{ViLBERT_Lu_ViLBERTPretraining2019} introduced a two-stream Transformer with co-attentional layers pre-trained on image--text pairs, while UNITER~\cite{UNITER_Chen_UNITERUniversal2020} adopted a single-stream Transformer to learn universal image--text representations via masked language/region modeling and image--text matching.
These pre-trained multimodal Transformers substantially improved downstream vision--language tasks but required costly cross-attention at inference time, limiting their use in large-scale retrieval.

Building on these foundations, a new generation of models scaled up contrastive dual-encoder pre-training to web-scale data.
Representative methods include CLIP~\cite{CLIP_Radford_LearningTransferable2021}, ALIGN~\cite{ALIGN_Jia_ScalingVisual2021}, BLIP~\cite{BLIP_Li_BLIPBootstrapping2022}, and SigLIP~\cite{SigLIP_Zhai_SigmoidLoss2023}, which encode images and text through independent encoders and align them in a shared representation space via contrastive learning.
While these models have achieved remarkable success across a wide range of vision--language tasks, their dual-encoder design inherently limits the degree of cross-modal interaction, making it difficult to capture fine-grained correspondences between visual and linguistic content.

With the rapid advancement of Multimodal Large Language Models (MLLMs), a new paradigm has emerged that repurposes MLLMs as embedding models.
VLM2Vec~\cite{VLM2Vec_Jiang_VLM2VecTraining2025} fine-tunes pre-trained MLLMs---such as Phi-3.5~\cite{Phi-3_Abdin_Phi3Technical2024}, LLaVA~\cite{LLaVA-NeXT-Interleave_Li_LLaVANeXTInterleaveTackling2024}, and Qwen2-VL~\cite{Qwen2-VL_Wang_Qwen2VLEnhancing2024}---on a large-scale multimodal embedding dataset, substantially outperforming conventional CLIP-based models on multimodal embedding benchmarks.
Building upon this, VLM2Vec-V2~\cite{VLM2Vec-V2_Meng_VLM2VecV2Advancing2025} further expands the training data to encompass not only images and text but also videos and visual documents, broadening the applicability of MLLM-based embeddings.
These methods demonstrate that the rich world knowledge and deep cross-modal understanding acquired during MLLM pre-training can be effectively distilled into dense vector representations.

Motivated by these advances, SIMMER applies MLLM-based embedding models to the cross-modal retrieval of food images and recipe texts---a domain in which only dual-encoder approaches have been explored to date.

\subsection{Cross-Modal Food Image--Recipe Retrieval}

Cross-modal retrieval between food images and recipe texts has become an actively studied task since the release of the Recipe1M dataset~\cite{Recipe1M_Salvador_LearningCrossModal2017}.
Compared to generic image--text retrieval, a key characteristic of this problem is that recipe texts are inherently structured, typically consisting of a title, a list of ingredients, and a sequence of cooking instructions.
Accordingly, progress in this area has been closely tied to (i) how each modality is encoded and (ii) how cross-modal alignment objectives exploit this structured text.

Early approaches largely followed a CNN--RNN recipe encoding pipeline: image encoders were typically CNNs pre-trained on ImageNet~\cite{ImageNet_Deng_ImageNetLargescale2009}, such as VGG~\cite{VGG_Simonyan_VeryDeep2014} or ResNet~\cite{ResNet_He_DeepResidual2016}, while recipe encoders used Word2Vec~\cite{Word2Vec_Mikolov_EfficientEstimation2013} embeddings followed by LSTMs~\cite{LSTM_Hochreiter_LongShortTerm1997}.
Within this paradigm, Salvador et al.~\cite{Recipe1M_Salvador_LearningCrossModal2017} introduced a joint embedding model trained with a cosine-similarity-based ranking objective augmented by semantic regularization.
AdaMine~\cite{Adamine_Carvalho_CrossModalRetrieval2018} proposed a joint retrieval-and-classification framework with a double-triplet learning scheme and adaptive triplet mining.
Several subsequent works improved alignment or robustness by adding adversarial and generative auxiliary objectives~\cite{ACME_Wang_LearningCrossModal2019,R2GAN_Zhu_R2GANCrossModal2019}, modeling cooking procedures more explicitly~\cite{Chen_DeepUnderstanding2018}, introducing noise-robust and sentence-level recipe modeling~\cite{SN_Zan_SentencebasedNoiserobust2020}, or incorporating stochastic latent variables to capture cross-modal interactions while keeping inference efficient~\cite{MCEN_Fu_MCENBridging2020}.
While they were effective, these early recipe encoders generally relied on limited language pre-training (e.g., Word2Vec) and struggled to capture richer compositional semantics of long, structured recipes.

A second wave of work emphasized richer recipe modeling and finer-grained image--recipe correspondences.
This includes structure-aware recipe encoders, such as tree-structured LSTMs that capture hierarchical relationships among recipe components~\cite{CHEF_Pham_CHEFCrossmodal2021}, as well as Transformer-based multilingual recipe encoders with translation-based regularization as in X-MRS~\cite{X-MRS_Guerrero_CrossmodalRetrieval2021}.
In parallel, many methods focused on designing stronger cross-modal interaction and alignment mechanisms (e.g., hybrid fusion and intra-/inter-modality attention~\cite{HF-ICMA_Li_HybridFusion2021,IMHF_Li_CrossModalImageRecipe2021}, multi-subspace implicit alignment~\cite{M-SIA_Li_MultisubspaceImplicit2021}, modality-alignment regularizations and improved metric learning objectives~\cite{JEMA_Xie_LearningJoint2021,SEJE_Xie_LearningTextimage2021}, semantic consistency with attention for discriminative recipe representations~\cite{SCAN_Wang_CrossModalFood2022}, and disentanglement-based generation as an auxiliary signal~\cite{RDE-GAN_Sugiyama_CrossModalRecipe2021}).

With the rise of the Transformer architecture~\cite{Transformer_Vaswani_AttentionAll2017}, stronger backbones became increasingly common on both modalities.
On the recipe side, the Hierarchical Transformer (H-T)~\cite{H-T_Salvador_RevampingCrossModal2021} became influential by encoding recipe components (title, ingredients, instructions) and introducing a self-supervised objective over recipe components, enabling the use of both paired image--recipe data and recipe-only samples.
On top of such modern recipe encoders, some methods explicitly model cross-modal interactions during training while retaining efficient unimodal encoding at inference, e.g., Transformer decoders used as a multimodal regularizer in T-Food~\cite{T-Food_Shukor_TransformerDecoders2022}.
Around the same time, other complementary directions were explored: non-parametric retrieval on top of precomputed embeddings with cross-modal kNN-type alignment (DaC)~\cite{DaC_Fain_DividingConquering2019}, event-oriented modality alignment for event-dense text--image retrieval validated in the food domain (EOMA)~\cite{EOMA_Xie_CrossModalRetrieval2022}, TF-IDF-enhanced recipe representations for joint embedding learning (MSJE)~\cite{MSJE_Xie_LearningTFIDF2022}, and structured “cooking program” representations to enrich procedural semantics~\cite{Papadopoulos_LearningProgram2022}.
More recent Transformer-based pipelines further refine recipe encoders and training regimes (e.g., large-batch training and improved component-level learning)~\cite{TNLBT_Yang_TransformerBasedCrossModal2023,TNLBT2_Yang_ImprovingCrossModal2024}.

In recent years, large-scale vision--language pre-training, most notably CLIP~\cite{CLIP_Radford_LearningTransferable2021}, has become a powerful backbone for food retrieval when adapted to structured recipes.
Representative examples include component-aware prompt learning to leverage vision--language pre-trained models without full fine-tuning~\cite{CIP_Huang_ImprovingCrossModal2023}, adaptation of vision--language pre-training to structured recipe text~\cite{VLPCook_Shukor_VisionStructuredLanguage2024}, and foundation-model-driven data augmentation with lightweight adaptation (e.g., adapters) and multi-level alignment losses~\cite{DAR_Song_EnhancingRecipe2024}.
CLIP-style retrievers are also used as components in retrieval-augmented recipe generation frameworks, where retrieval quality directly affects generation quality~\cite{RecipeRAG_Yang_RecipeRAGAdvancing2025}.
In addition, many recent methods continue to strengthen alignment through task-specific designs such as fine-grained component alignment~\cite{FARM_Wahed_FineGrainedAlignment2024}, disambiguation and partial-matching objectives~\cite{MMACMR_Zou_DisambiguityAlignment2024,CREAMY_Zou_CREAMYCrossModal2024}, unified text encoding for efficiency~\cite{UTE-FCL_Zhang_CrossmodalRecipe2024}, improved visual--textual interaction modules~\cite{FMI_Zhao_CrossModal2025,DCA-Food_Liu_RevampingImageRecipe2024}, and mitigation of representation bias due to missing visual evidence in food images~\cite{Wang_MitigatingCrossmodal2025}.

Despite this steady progression, most prior work is designed for scalable large-scale retrieval and is typically deployed as a dual-encoder system at inference time, where food images and recipes are encoded separately into a shared embedding space (even if additional cross-modal modules are used during training or for re-ranking).
In contrast, SIMMER departs from this convention by employing a single MLLM-based embedding model as a unified encoder shared across both modalities.

\section{Method}
\subsection{Overview}

Conventional approaches to cross-modal food retrieval adopt a dual-encoder architecture, in which a visual encoder and a recipe text encoder are independently designed and trained to project their respective inputs into a shared embedding space.
In contrast, SIMMER employs a single MLLM as a unified encoder for both food images and recipe texts, mapping both modalities into the same embedding space through a shared model.
This design allows us to leverage the broad visual and linguistic knowledge that the MLLM has acquired through large-scale pre-training.

\subsection{Problem Formulation}
\label{sec:formulation}

Given a training set of $N$ image--recipe pairs $\{(v_i, r_i)\}_{i=1}^{N}$, where $v_i$ is a food image and $r_i$ is the corresponding recipe text, our goal is to learn a unified encoder $\Phi(\cdot)$ that maps both modalities into a common embedding space.
Each recipe $r_i$ is a structured tuple consisting of a title, a list of ingredients, and a sequence of cooking instructions.
We seek to learn $\Phi$ such that the image embedding $\mathbf{e}_v = \Phi(v)$ and the recipe embedding $\mathbf{e}_r = \Phi(r)$ are close for matching pairs and far apart for non-matching ones.
In our framework, $\Phi(\cdot)$ is instantiated as a Multimodal Large Language Model (MLLM) that natively accepts both visual and textual inputs.

\subsection{Unified Encoding via MLLM}

Following VLM2Vec~\cite{VLM2Vec_Jiang_VLM2VecTraining2025} and VLM2Vec-V2~\cite{VLM2Vec-V2_Meng_VLM2VecV2Advancing2025}, we obtain the embedding for a given input by extracting the hidden-state vector of the last token from the final layer of the MLLM.

In the VLM2Vec framework, contrastive learning is formulated as a unidirectional retrieval task from queries to candidates, rather than the symmetric formulation commonly used in standard contrastive learning.
Consequently, the same data point receives different prompt templates depending on whether it serves as a query or a candidate: queries are accompanied by search-oriented instructions, while candidates are given simpler prompts that elicit a general embedding representation.

Following this design, we define two prompt templates for each modality---one for the query role and one for the candidate role---corresponding to the two retrieval directions: image-to-recipe and recipe-to-image.

\subsubsection{Visual Embedding Generation}

Listings~\ref{lst:i2r_img_query} and~\ref{lst:r2i_img_candidate} show the prompt templates used for encoding food images.
The placeholder \verb!<|image_1|>! denotes the position where the image is inserted.
When the image serves as a query in image-to-recipe retrieval, the prompt instructs the model to find a recipe corresponding to the given food image.
When the image serves as a candidate in recipe-to-image retrieval, the prompt instead directs the model to produce a visual representation suitable for recipe matching.

\noindent
\begin{minipage}{\linewidth}
\begin{lstlisting}[caption={
    Prompt for Visual Query in Image-to-Recipe Retrieval
    }, label={lst:i2r_img_query}]
<|image_1|>
Find a cooking recipe describing the given food image.
\end{lstlisting}

\begin{lstlisting}[caption={
    Prompt for Visual Candidate in Recipe-to-Image Retrieval
    }, label={lst:r2i_img_candidate}]
<|image_1|>
Represent the given food image for recipe prediction.
\end{lstlisting}
\end{minipage}

\subsubsection{Recipe Text Embedding Generation}

Listings~\ref{lst:r2i_recipe_query} and~\ref{lst:i2r_recipe_candidate} show the prompt templates for encoding recipe texts.
In the templates, \verb|{title}| denotes the recipe title, \verb|{ingredients}| is the comma-separated list of ingredients, and \verb|{instructions}| is the sequence of cooking steps joined by spaces.
When the recipe serves as a query in recipe-to-image retrieval, the prompt instructs the model to find a food image matching the given recipe.
When the recipe serves as a candidate in image-to-recipe retrieval, a simpler descriptive prefix is used to elicit a general recipe embedding.

\noindent
\begin{minipage}{\linewidth}
\begin{lstlisting}[caption={
    Prompt for Recipe Query in Recipe-to-Image Retrieval
    }, label={lst:r2i_recipe_query}]
Find me a food image that matches the given cooking recipe: Title: {title}, Ingredients: {ingredients}, Instructions: {instructions}
\end{lstlisting}

\begin{lstlisting}[caption={
    Prompt for Recipe Candidate in Image-to-Recipe Retrieval
    }, label={lst:i2r_recipe_candidate}]
A cooking recipe: Title: {title}, Ingredients: {ingredients}, Instructions: {instructions}
\end{lstlisting}
\end{minipage}

\subsection{Component-Aware Data Augmentation}
\label{sec:augmentation}

In real-world scenarios, recipe data may be incomplete---for example, a recipe might lack a title or an explicit ingredient list.
To improve the robustness of the model to such partial inputs, we propose a component-aware data augmentation strategy.
Specifically, we generate augmented training samples by removing two of the three recipe components, retaining only a single component.
This yields three additional variants for each original recipe, resulting in a total of four training patterns per image--recipe pair:
\begin{enumerate}
    \item The complete recipe containing the title, ingredients, and instructions.
    \item Title only.
    \item Ingredients only.
    \item Instructions only.
\end{enumerate}
By training on both complete and partial recipes, the model learns to produce informative embeddings even when only a subset of recipe information is available.
We evaluate the effectiveness of this strategy in Section~\ref{sec:recipe-components}.

\subsection{Training Objectives}

\begin{figure}
  \centering
  \includegraphics[width=\linewidth]{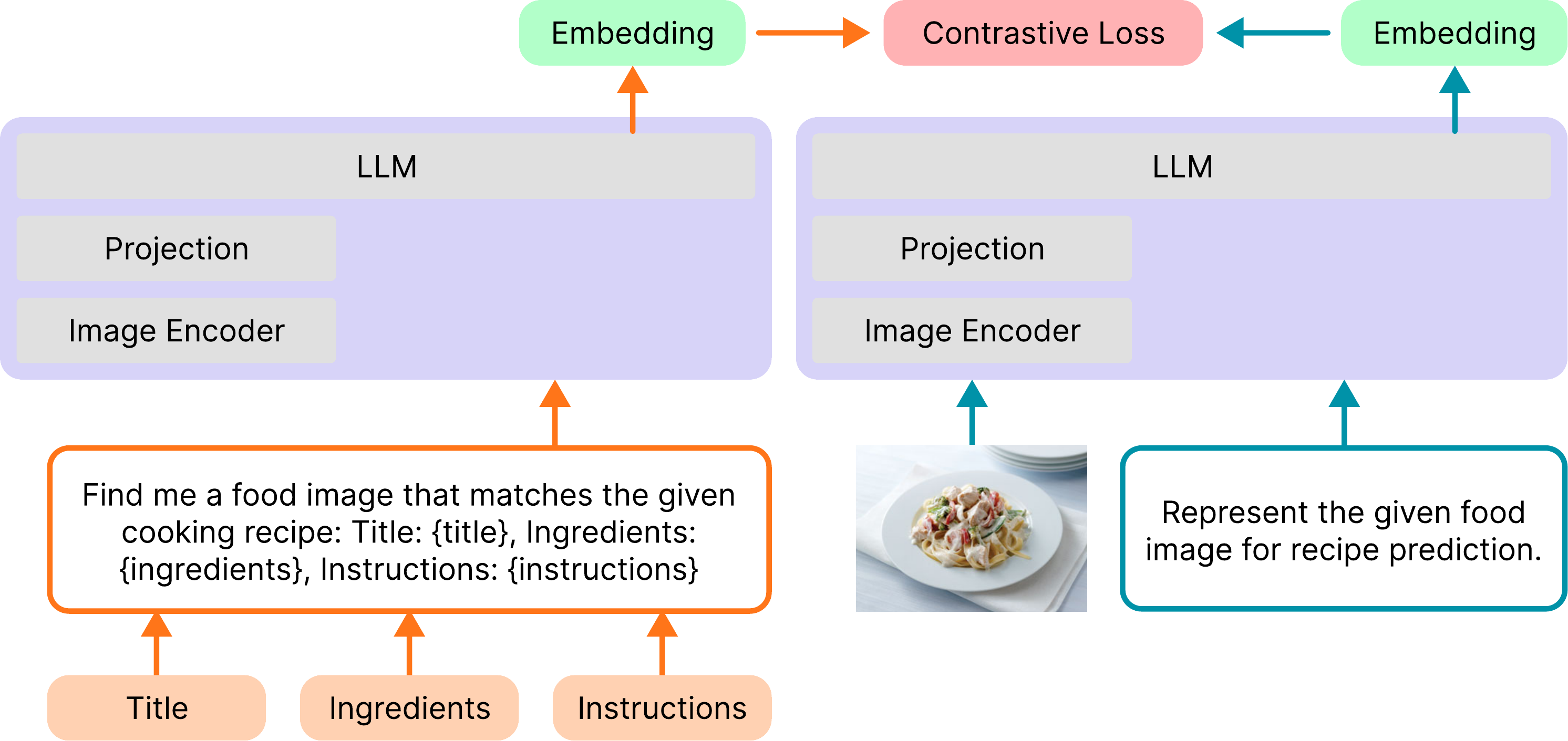}
  \caption{Overview of the contrastive learning procedure used for fine-tuning}
  \label{fig:contrastive}
\end{figure}

To achieve the goal described in Section~\ref{sec:formulation}, we fine-tune the MLLM via contrastive learning following the VLM2Vec framework, as illustrated in Figure~\ref{fig:contrastive}.
Since VLM2Vec formulates learning as a unidirectional retrieval task from queries to candidates, we construct two separate datasets from the same set of image--recipe pairs: one for image-to-recipe retrieval and one for recipe-to-image retrieval.
A unidirectional information noise contrastive estimation (InfoNCE) loss~\cite{InfoNCE_Oord_RepresentationLearning2018} is computed for each dataset, thereby enabling bidirectional retrieval through the combination of both directions.

Concretely, given a mini-batch of $B$ query--candidate pairs $\{(q_i, c_i)\}_{i=1}^{B}$, the matched pair is treated as the positive and all other in-batch candidates serve as negatives.
The InfoNCE loss is defined as:
\begin{equation}
    \label{eq:info_nce_loss}
    \mathcal{L} = -\frac{1}{B} \sum_{i=1}^{B} \log \frac{\exp(\mathrm{sim}(\mathbf{e}_{q_i}, \mathbf{e}_{c_i}) / \tau)}{\sum_{j=1}^{B} \exp(\mathrm{sim}(\mathbf{e}_{q_i}, \mathbf{e}_{c_j}) / \tau)}
\end{equation}
where $\mathbf{e}_{q_i}$ and $\mathbf{e}_{c_i}$ are the embeddings of the query and the candidate, respectively, $\mathrm{sim}(\cdot, \cdot)$ denotes cosine similarity, and $\tau$ is a temperature hyperparameter.

\section{Experiments}
\subsection{Experimental Setup}
\subsubsection{Dataset}

Following prior work, we evaluate SIMMER on the Recipe1M dataset~\cite{Recipe1M_Salvador_LearningCrossModal2017}, a large-scale collection of over one million cooking recipes paired with food images.
Since not all recipes are associated with images, we use only the subset with available image--recipe pairs, resulting in 238,408 pairs for training, 51,119 for validation, and 51,304 for testing.
Each recipe consists of a title string, a list of ingredient strings, and a list of cooking instruction strings.

\subsubsection{Evaluation Metrics}

Following prior work, we adopt median rank (medR) and Recall@$k$ (R@$k$, $k = 1, 5, 10$) as evaluation metrics.
The median rank measures the median position of the ground-truth item in the ranked retrieval list, where lower values indicate better performance.
Recall@$k$ measures the proportion of queries for which the ground-truth item appears within the top-$k$ retrieved results.

We report results under two standard evaluation settings: the 1k setup and the 10k setup.
In the 1k setup, 1,000 test pairs are randomly sampled and metrics are computed over this subset; this procedure is repeated 10 times and the results are averaged to mitigate the variance introduced by random sampling.
The 10k setup increases the candidate pool to 10,000 pairs, making the retrieval task substantially more challenging.
To ensure a fair comparison, we compute all metrics using the publicly available evaluation code provided by Salvador et al.~\cite{H-T_Salvador_RevampingCrossModal2021}\footnote{\url{https://github.com/amzn/image-to-recipe-transformers}}.

\subsubsection{Implementation Details}

All experiments are conducted on 8 NVIDIA RTX A6000 GPUs.
We adopt three models from the VLM2Vec family as base models; their specifications are summarized in Table~\ref{tab:base_models}.

We fine-tune each model using the Adam optimizer~\cite{Adam_Kingma_AdamMethod2015} with a learning rate of $1 \times 10^{-4}$ for 2,000 steps.
The training objective is the InfoNCE loss with a temperature parameter of $\tau = 0.02$.
The global batch size is set to 128, and we employ GradCache~\cite{GradCache_Gao_ScalingDeep2021} to enable large-batch contrastive learning under limited GPU memory.
For parameter-efficient fine-tuning, we apply Low-Rank Adaptation (LoRA)~\cite{LoRA_Hu_LoRALowRank2022} with a rank of 16, a scaling factor ($\alpha$) of 64, and a dropout rate of 0.1.
The total training time is approximately 20 hours per model.

\begin{table}[t]
  \centering
  \begin{threeparttable}
    \caption{Base models used in our experiments}
    \label{tab:base_models}

    \begin{tabular*}{\linewidth}{@{\extracolsep{\fill}}lcc}
      \toprule
      Model & Parameters & Embedding Dimension \\
      \midrule
      VLM2Vec-V1-2B\tnote{1} & 2B & 1,536 \\
      VLM2Vec-V1-7B\tnote{2} & 7B & 3,584 \\
      VLM2Vec-V2\tnote{3}    & 2B & 1,536 \\
      \bottomrule
    \end{tabular*}

    \begin{tablenotes}
      \footnotesize
      \item[1] \url{https://huggingface.co/TIGER-Lab/VLM2Vec-Qwen2VL-2B}
      \item[2] \url{https://huggingface.co/TIGER-Lab/VLM2Vec-Qwen2VL-7B}
      \item[3] \url{https://huggingface.co/VLM2Vec/VLM2Vec-V2.0}
    \end{tablenotes}
  \end{threeparttable}
\end{table}

\subsection{Comparison with State-of-the-Art Methods}

Table~\ref{tab:recipe1m} presents the comparison of SIMMER with existing approaches on the Recipe1M dataset.
We compare against a comprehensive set of state-of-the-art methods spanning the full history of this task; detailed descriptions of all baseline methods are provided in Appendix~\ref{sec:baselines}.
All three variants of SIMMER---SIMMER (V1-2B), SIMMER (V1-7B), and SIMMER (V2)---consistently outperform all prior methods across both the 1k and 10k evaluation settings.
In particular, SIMMER (V1-7B) achieves the best overall performance, attaining R@1 scores of 87.5\% and 85.1\% on the 1k image-to-recipe and recipe-to-image tasks, respectively, and 65.5\% and 61.5\% on the corresponding 10k tasks.
Compared to the previous best method, Yang et al.~\cite{TNLBT2_Yang_ImprovingCrossModal2024}, SIMMER (V1-7B) improves the 1k image-to-recipe R@1 from 81.8\% to 87.5\% and the 10k image-to-recipe R@1 from 56.5\% to 65.5\%, representing substantial absolute gains of 5.7 and 9.0 percentage points, respectively.

An interesting observation is that SIMMER (V1-7B) outperforms SIMMER (V2) despite VLM2Vec-V2 being a more recent model.
Note that both VLM2Vec-V1 and VLM2Vec-V2 share the same underlying backbone architecture, Qwen2-VL, which allows us to isolate the factors contributing to this difference.
We suspect that two factors account for the performance gap.
First, VLM2Vec-V2 is trained on a broader range of data modalities, including video and visual document understanding tasks, in addition to image--text data.
However, these additional training signals do not appear to transfer effectively to the food image--recipe retrieval task.
Second, and more importantly, VLM2Vec-V2 is available only as a 2B-parameter model with 1,536-dimensional embeddings, whereas VLM2Vec-V1 offers a 7B-parameter variant that produces 3,584-dimensional embeddings.
The superior performance of V1-7B over both V1-2B and V2 suggests that model capacity and embedding dimensionality play a critical role in capturing the fine-grained cross-modal correspondences required for recipe retrieval.

\begin{table*}[t]
\caption{Experimental results on image-to-recipe and recipe-to-image retrieval. The best results are presented in \textbf{bold}, while the next best results are presented in \uline{underlined}. A full comparison with all baseline methods is provided in Tables~\ref{tab:recipe1m-full-1k} and~\ref{tab:recipe1m-full-10k}. SIMMER denotes our proposed method}
\label{tab:recipe1m}
\begin{adjustbox}{max width=\textwidth}
\begin{tabular}{clccccccccc}
\toprule
\multirow{2}{*}{Setting} & \multirow{2}{*}{Method} & \multirow{2}{*}{Venue} & \multicolumn{4}{c}{Image-to-Recipe} & \multicolumn{4}{c}{Recipe-to-Image} \\
\cmidrule(lr){4-7}\cmidrule(lr){8-11}
                      &                         &                        & medR\,$\downarrow$ & R@1\,$\uparrow$ & R@5\,$\uparrow$ & R@10\,$\uparrow$ & medR\,$\downarrow$ & R@1\,$\uparrow$ & R@5\,$\uparrow$ & R@10\,$\uparrow$ \\
\midrule
\multirow{22}{*}{1k}
& Salvador et al.~\cite{Recipe1M_Salvador_LearningCrossModal2017} & CVPR'17                 & 5.2  & 24.0          & 51.0          & 65.0          & 5.1  & 25.0          & 52.0          & 65.0          \\
& AdaMine~\cite{Adamine_Carvalho_CrossModalRetrieval2018}         & SIGIR'18                & 1.0  & 39.8          & 69.0          & 77.7          & 1.0  & 40.2          & 68.1          & 78.7          \\
& ACME~\cite{ACME_Wang_LearningCrossModal2019}                    & CVPR'19                 & 1.0  & 51.8          & 80.2          & 87.5          & 1.0  & 52.8          & 80.2          & 87.6          \\
& MCEN~\cite{MCEN_Fu_MCENBridging2020}                            & CVPR'20                 & 2.0  & 48.2          & 75.8          & 83.6          & 1.9  & 48.4          & 76.1          & 83.7          \\
& X-MRS~\cite{X-MRS_Guerrero_CrossmodalRetrieval2021}             & MM'21                   & 1.0  & 64.0          & 88.3          & 92.6          & 1.0  & 63.9          & 87.6          & 92.6          \\
& RDE-GAN~\cite{RDE-GAN_Sugiyama_CrossModalRecipe2021}            & MM'21                   & 1.0  & 59.4          & 81.0          & 87.4          & 1.0  & 61.2          & 81.0          & 87.2          \\
& H-T~\cite{H-T_Salvador_RevampingCrossModal2021}                 & CVPR'21                 & 1.0  & 60.0          & 87.6          & 92.9          & 1.0  & 60.3          & 87.6          & 93.2          \\
& T-Food~\cite{T-Food_Shukor_TransformerDecoders2022}             & CVPRW'22                & 1.0  & 72.3          & 90.7          & 93.4          & 1.0  & 72.6          & 90.6          & 93.4          \\
& TNLBT~\cite{TNLBT_Yang_TransformerBasedCrossModal2023}          & MMM'23                  & 1.0  & 81.0          & 95.2          & 97.4          & 1.0  & 80.3          & 95.2          & 97.4          \\
& CIP~\cite{CIP_Huang_ImprovingCrossModal2023}                    & MM'23                   & 1.0  & 77.1          & 94.2          & 97.2          & 1.0  & 77.3          & 94.4          & 97.0          \\
& CREAMY~\cite{CREAMY_Zou_CREAMYCrossModal2024}                   & IEEE Access'24         & 1.0  & 73.3          & 92.5          & 95.6          & 1.0  & 73.2          & 92.5          & 95.8          \\
& Yang et al.~\cite{TNLBT2_Yang_ImprovingCrossModal2024}          & ICMRW'24                & 1.0  & \uline{81.8}  & \uline{95.9}  & \uline{97.8}  & 1.0  & \uline{81.2}  & \uline{96.0}  & \uline{97.9}  \\
& VLPCook~\cite{VLPCook_Shukor_VisionStructuredLanguage2024}      & CVIU'24                & 1.0  & 73.6          & 90.5          & 93.3          & 1.0  & 74.7          & 90.7          & 93.2          \\
& FARM~\cite{FARM_Wahed_FineGrainedAlignment2024}                 & WACV'24                 & 1.0  & 73.7          & 90.7          & 93.4          & 1.0  & 73.6          & 90.8          & 93.5          \\
& UTE-FCL~\cite{UTE-FCL_Zhang_CrossmodalRecipe2024}               & Knowledge-Based Systems'24& 1.0  & 68.1          & 90.0          & 94.5          & 1.0  & 67.2          & 90.1          & 94.7          \\
& DAR~\cite{DAR_Song_EnhancingRecipe2024}                         & ECCV'24                 & 1.0  & 77.3          & 95.3          & 97.7          & 1.0  & 77.1          & 95.4          & \uline{97.9}  \\
& FMI~\cite{FMI_Zhao_CrossModal2025}                              & Scientific Reports'25  & 1.0  & 77.4          & 95.8          & 97.6          & 1.0  & 77.1          & 95.4          & 97.7          \\
& Wang et al.~\cite{Wang_MitigatingCrossmodal2025}                & MM'25                   & 1.0  & 79.1          & 94.6          & 97.0          & 1.0  & 78.3          & 95.0          & 97.2          \\
\cmidrule(lr){2-11}
& SIMMER (V1-2B)                                                    & -                       & 1.0  & 84.1          & 97.3          & 98.8          & 1.0  & 81.5          & 96.6          & 98.4          \\
& SIMMER (V1-7B)                                                    & -                       & 1.0  & \textbf{87.5} & \textbf{98.0} & \textbf{99.2} & 1.0  & \textbf{85.1} & \textbf{97.6} & \textbf{99.1} \\
& SIMMER (V2)                                                       & -                       & 1.0  & 83.8          & 96.9          & 98.7          & 1.0  & 81.7          & 96.5          & 98.3          \\
\midrule
\multirow{22}{*}{10k}
& Salvador et al.~\cite{Recipe1M_Salvador_LearningCrossModal2017} & CVPR'17                 & 41.9 & -             & -             & -             & 39.2 & -             & -             & -             \\
& AdaMine~\cite{Adamine_Carvalho_CrossModalRetrieval2018}         & SIGIR'18                & 13.2 & 14.9          & 35.3          & 45.2          & 12.2 & 14.8          & 34.6          & 46.1          \\
& ACME~\cite{ACME_Wang_LearningCrossModal2019}                    & CVPR'19                 & 6.7  & 22.9          & 46.8          & 57.9          & 6.0  & 24.4          & 47.9          & 59.0          \\
& MCEN~\cite{MCEN_Fu_MCENBridging2020}                            & CVPR'20                 & 7.2  & 20.3          & 43.3          & 54.4          & 6.6  & 21.4          & 44.3          & 55.2          \\
& X-MRS~\cite{X-MRS_Guerrero_CrossmodalRetrieval2021}             & MM'21                   & 3.0  & 32.9          & 60.6          & 71.2          & 3.0  & 33.0          & 60.4          & 70.7          \\
& RDE-GAN~\cite{RDE-GAN_Sugiyama_CrossModalRecipe2021}            & MM'21                   & 3.5  & 36.0          & 56.1          & 64.4          & 3.0  & 38.2          & 57.7          & 65.8          \\
& H-T~\cite{H-T_Salvador_RevampingCrossModal2021}                 & CVPR'21                 & 4.0  & 27.9          & 56.4          & 68.1          & 4.0  & 28.3          & 56.5          & 68.1          \\
& T-Food~\cite{T-Food_Shukor_TransformerDecoders2022}             & CVPRW'22                & 2.0  & 43.4          & 70.7          & 79.7          & 2.0  & 44.6          & 71.2          & 79.7          \\
& TNLBT~\cite{TNLBT_Yang_TransformerBasedCrossModal2023}          & MMM'23                  & 1.0  & \uline{56.5}  & 80.7          & 87.1          & 1.0  & \uline{55.9}  & 80.1          & 86.8          \\
& CIP~\cite{CIP_Huang_ImprovingCrossModal2023}                    & MM'23                   & 2.0  & 44.9          & 72.8          & 82.0          & 2.0  & 45.2          & 73.0          & 81.8          \\
& CREAMY~\cite{CREAMY_Zou_CREAMYCrossModal2024}                   & IEEE Access'24         & 2.0  & 44.6          & 71.6          & 80.4          & 2.0  & 45.0          & 71.4          & 80.0          \\
& Yang et al.~\cite{TNLBT2_Yang_ImprovingCrossModal2024}          & ICMRW'24                & 1.0  & \uline{56.5}  & \uline{81.0}  & \uline{87.6}  & 1.0  & 55.7          & \uline{80.2}  & \uline{87.1}  \\
& VLPCook~\cite{VLPCook_Shukor_VisionStructuredLanguage2024}      & CVIU'24                & 2.0  & 45.3          & 72.4          & 80.8          & 2.0  & 46.4          & 73.1          & 80.9          \\
& FARM~\cite{FARM_Wahed_FineGrainedAlignment2024}                 & WACV'24                 & 2.0  & 44.9          & 71.8          & 80.0          & 2.0  & 44.3          & 71.5          & 80.0          \\
& UTE-FCL~\cite{UTE-FCL_Zhang_CrossmodalRecipe2024}               & Knowledge-Based Systems'24& 2.7  & 37.4          & 65.4          & 75.4          & 3.0  & 36.5          & 64.7          & 74.8          \\
& DAR~\cite{DAR_Song_EnhancingRecipe2024}                         & ECCV'24                 & 2.0  & 47.8          & 75.9          & 84.3          & 2.0  & 47.4          & 75.5          & 84.1          \\
& FMI~\cite{FMI_Zhao_CrossModal2025}                              & Scientific Reports'25  & 1.0  & 48.4          & 76.3          & 81.9          & 1.0  & 49.5          & 79.2          & 83.1          \\
& Wang et al.~\cite{Wang_MitigatingCrossmodal2025}                & MM'25                   & 1.0  & 51.7          & 78.2          & 85.9          & 1.0  & 52.2          & 78.4          & 86.0          \\
\cmidrule(lr){2-11}
& SIMMER (V1-2B)                                                    & -                       & 1.0  & 59.7          & 83.5          & 89.9          & 1.0  & 55.8          & 81.1          & 88.0          \\
& SIMMER (V1-7B)                                                    & -                       & 1.0  & \textbf{65.5} & \textbf{87.4} & \textbf{92.5} & 1.0  & \textbf{61.5} & \textbf{85.0} & \textbf{91.0} \\
& SIMMER (V2)                                                       & -                       & 1.0  & 59.1          & 83.3          & 89.8          & 1.0  & 55.7          & 81.0          & 88.0          \\
\bottomrule
\end{tabular}
\end{adjustbox}
\end{table*}

\subsection{Ablation Study}

\subsubsection{Effect of Fine-tuning}

To validate the effectiveness of fine-tuning on the Recipe1M dataset, we compare our fine-tuned models against their zero-shot counterparts, i.e., the pre-trained VLM2Vec models applied directly without any task-specific training.
The results are summarized in Table~\ref{tab:fine-tuning}.
Across all three model variants, fine-tuning yields substantial performance improvements.
For instance, the V1-7B model achieves a 1k image-to-recipe R@1 of only 40.8\% in the zero-shot setting, whereas fine-tuning on Recipe1M boosts this to 87.5\%, an absolute gain of 46.7 percentage points.
Similar trends are observed for the other variants: V1-2B improves from 29.5\% to 84.1\% and V2 from 45.1\% to 83.8\% in the same metric.
The median rank also drops consistently to 1.0 across all fine-tuned configurations, representing a significant improvement over the zero-shot baselines.
These results confirm that, although VLM2Vec models possess general-purpose vision--language understanding capabilities, task-specific fine-tuning is essential for achieving strong cross-modal retrieval performance in the food domain.

\begin{table*}[t]
\caption{Effect of fine-tuning. Comparison between zero-shot inference and SIMMER on Recipe1M}
\label{tab:fine-tuning}
\begin{adjustbox}{max width=\textwidth}
\begin{tabular}{clcccccccc}
\toprule
\multirow{2}{*}{Setting} & \multirow{2}{*}{Method} & \multicolumn{4}{c}{Image-to-Recipe} & \multicolumn{4}{c}{Recipe-to-Image} \\
\cmidrule(lr){3-6}\cmidrule(lr){7-10}
                      &                         & medR\,$\downarrow$ & R@1\,$\uparrow$ & R@5\,$\uparrow$ & R@10\,$\uparrow$ & medR\,$\downarrow$ & R@1\,$\uparrow$ & R@5\,$\uparrow$ & R@10\,$\uparrow$ \\
\midrule
\multirow{6}{*}{1k}
& V1-2B (Zero-shot) & 4.65 & 29.5          & 51.9          & 61.2          & 7.7  & 18.4          & 42.8          & 57.9          \\
& SIMMER (V1-2B)      & 1.0  & 84.1          & 97.3          & 98.8          & 1.0  & 81.5          & 96.6          & 98.4          \\
\cmidrule(lr){2-10}
& V1-7B (Zero-shot) & 2.1  & 40.8          & 66.5          & 76.0          & 2.0  & 39.8          & 69.0          & 79.0          \\
& SIMMER (V1-7B)      & 1.0  & 87.5          & 98.0          & 99.2          & 1.0  & 85.1          & 97.6          & 99.1          \\
\cmidrule(lr){2-10}
& V2 (Zero-shot)    & 2.0  & 45.1          & 73.4          & 82.4          & 2.0  & 47.3          & 74.7          & 83.3          \\
& SIMMER (V2)         & 1.0  & 83.8          & 96.9          & 98.7          & 1.0  & 81.7          & 96.5          & 98.3          \\
\midrule
\multirow{6}{*}{10k}
& V1-2B (Zero-shot) & 40.4 & 11.4          & 25.7          & 33.3          & 69.9 & 6.2           & 14.1          & 20.1          \\
& SIMMER (V1-2B)      & 1.0  & 59.7          & 83.5          & 89.9          & 1.0  & 55.8          & 81.1          & 88.0          \\
\cmidrule(lr){2-10}
& V1-7B (Zero-shot) & 14.9 & 17.6          & 36.0          & 45.0          & 14.5 & 15.8          & 34.5          & 44.7          \\
& SIMMER (V1-7B)      & 1.0  & 65.5          & 87.4          & 92.5          & 1.0  & 61.5          & 85.0          & 91.0          \\
\cmidrule(lr){2-10}
& V2 (Zero-shot)    & 10.1 & 20.0          & 40.6          & 50.7          & 9.1  & 22.3          & 42.7          & 52.5          \\
& SIMMER (V2)         & 1.0  & 59.1          & 83.3          & 89.8          & 1.0  & 55.7          & 81.0          & 88.0          \\
\bottomrule
\end{tabular}
\end{adjustbox}
\end{table*}

\subsubsection{Analysis on Recipe Components}
\label{sec:recipe-components}

In practical applications, not all components of a recipe are always available.
For example, a user may only have a recipe title, or the ingredient list may be absent.
In this section, we evaluate retrieval performance when only a subset of recipe components is provided at inference time, and further examine the effectiveness of the data augmentation strategy proposed in Section~\ref{sec:augmentation}.
We use the V1-7B model and compare two training conditions: without data augmentation (V1-7B w/o Aug.) and with data augmentation (V1-7B).
The results are presented in Table~\ref{tab:recipe-components}.

First, we analyze the performance differences across varying combinations of recipe components.
When relying on a single component, using only the cooking instructions yields the highest performance, achieving an R@1 of 74.2\% on the 1k image-to-recipe task.
This suggests that cooking instructions contain the most informative cues for capturing the correspondence between a recipe and its visual appearance.
In contrast, using only the title (44.4\%) or only the ingredients (38.8\%) results in a substantial decline in performance.
Among two-component combinations, pairing ingredients with instructions achieves the best result, with an R@1 of 85.9\% on the 1k image-to-recipe task---approaching the 87.5\% obtained when all three components are used.
As expected, using all three components simultaneously yields the highest overall performance.

Next, we examine the effect of data augmentation.
The proposed augmentation strategy yields notable performance gains, particularly when some recipe components are missing.
For instance, when querying with only the title, data augmentation improves the 1k image-to-recipe R@1 from 40.4\% to 44.4\%; when using only ingredients, the score increases from 34.9\% to 38.8\%.
Notably, although the augmented training data consist exclusively of single-component recipes (i.e., each augmented sample retains only one of the three recipe elements), an improvement is also observed for the two-component combination of title and ingredients, which rises from 44.8\% to 46.9\%.

Conversely, the benefit of data augmentation is marginal for combinations that include cooking instructions as well as when all components are present.
This suggests that when instructions---which carry rich semantic information---are available, the model already achieves high performance and does not benefit further from the additional robustness introduced by augmentation.
Overall, these results confirm that the proposed data augmentation is particularly effective in scenarios where only components with limited informational content are available.

\begin{table*}[t]
\caption{Retrieval performance using different combinations of recipe components}
\label{tab:recipe-components}
\begin{adjustbox}{max width=\textwidth}
\begin{tabular}{lcccclcccccccc}
\toprule
\multirow{3}{*}{Setting} & \multicolumn{3}{c}{\multirow{2}{*}{Recipe components at inference}} & & \multirow{3}{*}{Method} & \multicolumn{4}{c}{Image-to-Recipe}                                       & \multicolumn{4}{c}{Recipe-to-Image}                                       \\
\cmidrule(lr){7-10}\cmidrule(lr){11-14}
                      & \multicolumn{3}{c}{}                                                & &                         & medR\,$\downarrow$ & R@1\,$\uparrow$ & R@5\,$\uparrow$ & R@10\,$\uparrow$ & medR\,$\downarrow$ & R@1\,$\uparrow$ & R@5\,$\uparrow$ & R@10\,$\uparrow$ \\
\cmidrule(lr){2-4}
                      & Title                   & Ingredients                       & Instructions                       & &                         &                    &                 &                 &                  &                    &                 &                 &                  \\
\midrule
\multirow{14}{*}{1k}
& \multirow{2}{*}{\ding{51}} & \multirow{2}{*}{}          & \multirow{2}{*}{}          & & V1-7B w/o Aug.      & 2.0                & 40.4            & 71.7            & 81.7             & 2.0                & 40.0            & 70.9            & 80.9             \\
&                            &                            &                            & & V1-7B               & 2.0                & 44.4            & 75.6            & 84.5             & 2.0                & 41.5            & 72.8            & 82.6             \\
\cmidrule(lr){2-14}
& \multirow{2}{*}{}          & \multirow{2}{*}{\ding{51}} & \multirow{2}{*}{}          & & V1-7B w/o Aug.      & 3.0                & 34.9            & 62.8            & 74.8             & 1.9                & 47.4            & 75.0            & 83.5             \\
&                            &                            &                            & & V1-7B               & 2.3                & 38.8            & 68.0            & 79.7             & 1.2                & 52.4            & 78.9            & 86.6             \\
\cmidrule(lr){2-14}
& \multirow{2}{*}{}          & \multirow{2}{*}{}          & \multirow{2}{*}{\ding{51}} & & V1-7B w/o Aug.      & 1.0                & 74.0            & 91.6            & 95.0             & 1.0                & 70.8            & 90.3            & 94.2             \\
&                            &                            &                            & & V1-7B               & 1.0                & 74.2            & 91.6            & 95.3             & 1.0                & 70.9            & 90.0            & 94.3             \\
\cmidrule(lr){2-14}
& \multirow{2}{*}{\ding{51}} & \multirow{2}{*}{\ding{51}} & \multirow{2}{*}{}          & & V1-7B w/o Aug.      & 2.0                & 44.8            & 74.5            & 84.4             & 1.0                & 58.3            & 83.9            & 90.3             \\
&                            &                            &                            & & V1-7B               & 2.0                & 46.9            & 78.2            & 87.1             & 1.0                & 62.3            & 87.1            & 92.4             \\
\cmidrule(lr){2-14}
& \multirow{2}{*}{\ding{51}} & \multirow{2}{*}{}          & \multirow{2}{*}{\ding{51}} & & V1-7B w/o Aug.      & 1.0                & 80.2            & 95.7            & 97.9             & 1.0                & 77.5            & 94.8            & 97.6             \\
&                            &                            &                            & & V1-7B               & 1.0                & 80.6            & 95.6            & 98.0             & 1.0                & 77.8            & 95.1            & 97.6             \\
\cmidrule(lr){2-14}
& \multirow{2}{*}{}          & \multirow{2}{*}{\ding{51}} & \multirow{2}{*}{\ding{51}} & & V1-7B w/o Aug.      & 1.0                & 85.9            & 97.5            & 98.8             & 1.0                & 83.6            & 97.0            & 98.6             \\
&                            &                            &                            & & V1-7B               & 1.0                & 85.9            & 97.4            & 98.8             & 1.0                & 83.2            & 96.9            & 98.5             \\
\cmidrule(lr){2-14}
& \multirow{2}{*}{\ding{51}} & \multirow{2}{*}{\ding{51}} & \multirow{2}{*}{\ding{51}} & & V1-7B w/o Aug.      & 1.0                & 87.6            & 98.2            & 99.1             & 1.0                & 85.3            & 97.7            & 99.1             \\
&                            &                            &                            & & V1-7B               & 1.0                & 87.5            & 98.0            & 99.2             & 1.0                & 85.1            & 97.6            & 99.1             \\
\midrule
\multirow{14}{*}{10k}
& \multirow{2}{*}{\ding{51}} & \multirow{2}{*}{}          & \multirow{2}{*}{}          & & V1-7B w/o Aug.      & 13.1               & 13.6            & 34.0            & 45.9             & 13.5               & 14.9            & 34.7            & 45.7             \\
&                            &                            &                            & & V1-7B               & 10.0               & 16.1            & 38.9            & 51.0             & 12.3               & 15.2            & 36.0            & 47.4             \\
\cmidrule(lr){2-14}
& \multirow{2}{*}{}          & \multirow{2}{*}{\ding{51}} & \multirow{2}{*}{}          & & V1-7B w/o Aug.      & 21.1               & 15.1            & 30.1            & 39.2             & 8.9                & 22.4            & 42.6            & 52.5             \\
&                            &                            &                            & & V1-7B               & 15.5               & 17.2            & 33.8            & 43.7             & 6.2                & 25.8            & 47.8            & 57.9             \\
\cmidrule(lr){2-14}
& \multirow{2}{*}{}          & \multirow{2}{*}{}          & \multirow{2}{*}{\ding{51}} & & V1-7B w/o Aug.      & 2.0                & 48.9            & 72.3            & 79.9             & 2.0                & 44.0            & 68.8            & 77.1             \\
&                            &                            &                            & & V1-7B               & 2.0                & 48.9            & 72.5            & 80.1             & 2.0                & 43.7            & 68.4            & 77.0             \\
\cmidrule(lr){2-14}
& \multirow{2}{*}{\ding{51}} & \multirow{2}{*}{\ding{51}} & \multirow{2}{*}{}          & & V1-7B w/o Aug.      & 10.0               & 19.4            & 39.5            & 50.6             & 4.1                & 30.1            & 54.1            & 64.0             \\
&                            &                            &                            & & V1-7B               & 8.8                & 21.4            & 41.9            & 53.1             & 3.1                & 34.0            & 58.9            & 68.9             \\
\cmidrule(lr){2-14}
& \multirow{2}{*}{\ding{51}} & \multirow{2}{*}{}          & \multirow{2}{*}{\ding{51}} & & V1-7B w/o Aug.      & 1.0                & 55.1            & 79.0            & 86.2             & 1.3                & 50.2            & 75.8            & 83.8             \\
&                            &                            &                            & & V1-7B               & 1.0                & 54.8            & 79.3            & 86.5             & 1.3                & 50.4            & 76.0            & 84.2             \\
\cmidrule(lr){2-14}
& \multirow{2}{*}{}          & \multirow{2}{*}{\ding{51}} & \multirow{2}{*}{\ding{51}} & & V1-7B w/o Aug.      & 1.0                & 63.2            & 85.6            & 91.2             & 1.0                & 59.4            & 83.0            & 89.4             \\
&                            &                            &                            & & V1-7B               & 1.0                & 63.2            & 85.7            & 91.2             & 1.0                & 59.1            & 82.9            & 89.3             \\
\cmidrule(lr){2-14}
& \multirow{2}{*}{\ding{51}} & \multirow{2}{*}{\ding{51}} & \multirow{2}{*}{\ding{51}} & & V1-7B w/o Aug.      & 1.0                & 65.6            & 87.3            & 92.6             & 1.0                & 61.8            & 84.9            & 90.9             \\
&                            &                            &                            & & V1-7B               & 1.0                & 65.5            & 87.4            & 92.5             & 1.0                & 61.5            & 85.0            & 91.0            \\
\bottomrule
\end{tabular}
\end{adjustbox}
\end{table*}

\subsection{Qualitative Analysis}
\subsubsection{Retrieval Examples}

\begin{figure*}
  \centering
  \includegraphics[width=\linewidth]{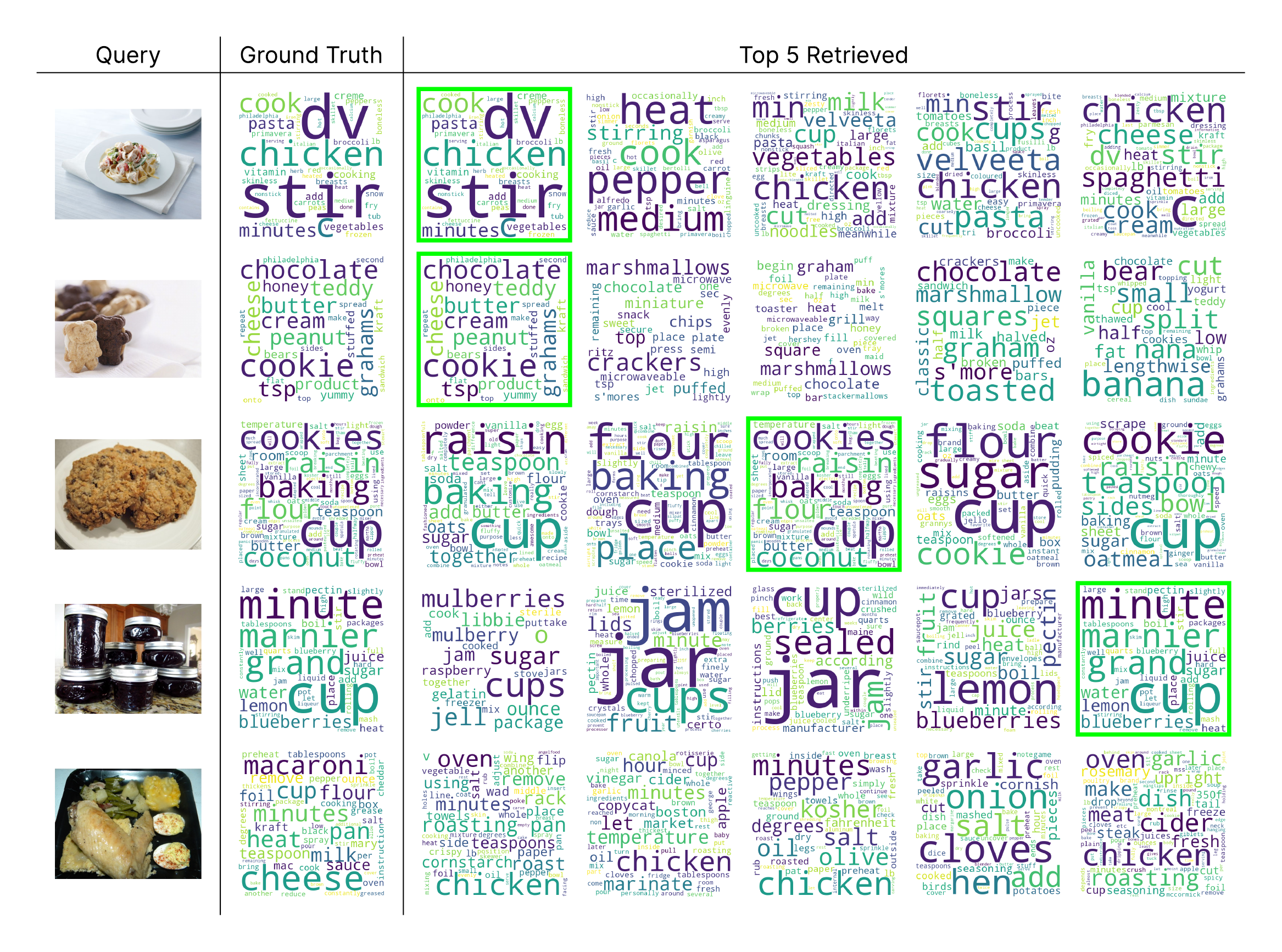}
  \caption{Qualitative examples of image-to-recipe retrieval by SIMMER (V1-7B). For each query image (left), the top-5 retrieved recipes are represented by word clouds, with the ground-truth recipe highlighted}
  \label{fig:i2r-search}
\end{figure*}

\begin{figure*}
  \centering
  \includegraphics[width=\linewidth]{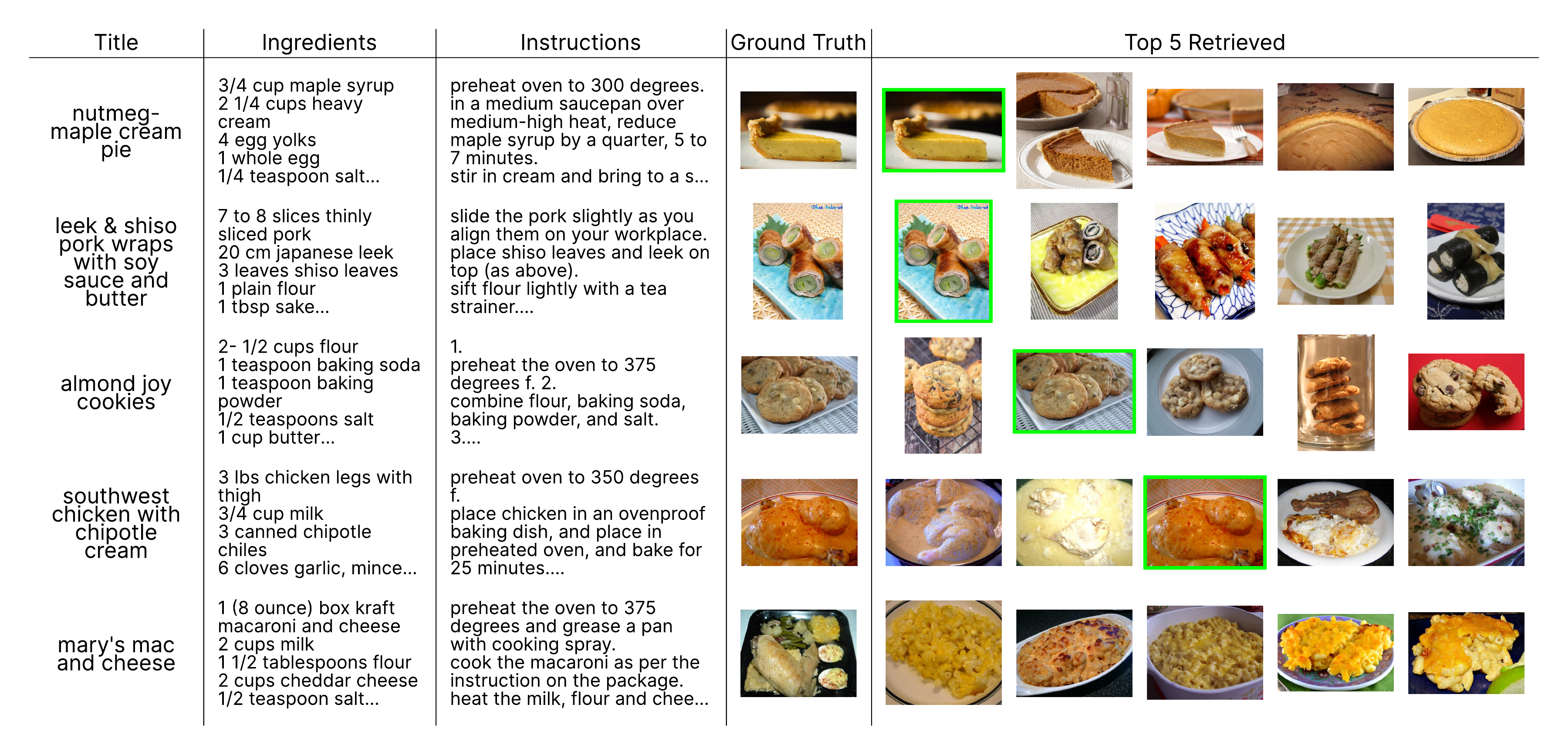}
  \caption{Qualitative examples of recipe-to-image retrieval by SIMMER (V1-7B). For each query recipe (left), the top-5 retrieved images are shown, with the ground-truth image highlighted}
  \label{fig:r2i-search}
\end{figure*}

We present qualitative examples of image-to-recipe and recipe-to-image retrieval by SIMMER (V1-7B) in Figures~\ref{fig:i2r-search} and~\ref{fig:r2i-search}, respectively.
In Figure~\ref{fig:i2r-search}, each recipe is visualized as a word cloud to provide an intuitive overview of its content.

Figure~\ref{fig:i2r-search} illustrates several image-to-recipe retrieval examples.
In the first example, the model correctly recognizes a pasta dish topped with chicken and vegetables from the query image and retrieves recipes that match this description as top-ranked results.
In the second example, the food image appears at first glance to depict ordinary cookies; however, the ground-truth recipe is a sandwich cookie with cream filling.
SIMMER successfully captures the cream that is faintly visible in the image and retrieves recipes related to s'mores\footnote{A traditional North American confection in which marshmallows and chocolate are sandwiched between graham crackers.} among the top results.
In particular, the ground-truth recipe is ranked first, likely because the model distinguishes the cookies in the image from the typical crackers used in s'mores.
The third example presents a challenging case where the query image lacks distinctive visual features and numerous visually similar oatmeal raisin cookie recipes exist in the database; nevertheless, the model retrieves the correct recipe at rank three.
The fourth example involves a blueberry jam whose image conveys little information beyond the appearance of a purple-colored jam, even to a human observer.
Despite this difficulty, SIMMER retrieves the correct recipe at rank five.
The last example highlights a problem inherent to the dataset rather than the model.
Although the dominant subject in the image is chicken, the annotated ground-truth recipe corresponds to the macaroni and cheese that appears only in the background.
Consequently, the model extracts features primarily from the prominent chicken, resulting in chicken-related recipes among the top results and the ground-truth recipe falling outside the top five.

Figure~\ref{fig:r2i-search} presents examples of recipe-to-image retrieval.
In the first example, visually similar pie images are retrieved in the top ranks, and the ground-truth image is correctly ranked first.
Notably, all lower-ranked images are of pumpkin pies, suggesting that the model captures the subtle visual differences between the ground-truth nutmeg-maple cream pie and the pumpkin pies.
In the second example, the query is a pork wrap recipe, and the top-ranked results include images of wraps and rolls.
The fifth-ranked image depicts a dish in which chicken is wrapped in kombu rather than pork; although its cooking instructions and visual appearance resemble those of the query, the title and ingredients differ substantially.
This indicates that the model correctly associates the wrapping action depicted in the images with the wrapping instructions described in the recipe, demonstrating its ability to capture fine-grained correspondences between visual and textual modalities.
In the third example, the query is a simple cookie recipe, and many visually similar cookie images exist in the dataset, preventing the model from placing the correct image at rank one.
The fourth example similarly involves numerous visually similar dish images, and the ground-truth image is retrieved at rank three.
The last example corresponds to the same data instance discussed in the bottom row of Figure~\ref{fig:i2r-search}.
Although the recipe describes macaroni and cheese, the annotated ground-truth image prominently features chicken, and consequently the correct image does not appear among the top five results.
This failure is attributable to a data mismatch rather than a model limitation, as the model successfully retrieves macaroni and cheese images as top-ranked results for the given recipe query.

\subsubsection{Qualitative Comparison}

\begin{figure*}
  \centering
  \includegraphics[width=\linewidth]{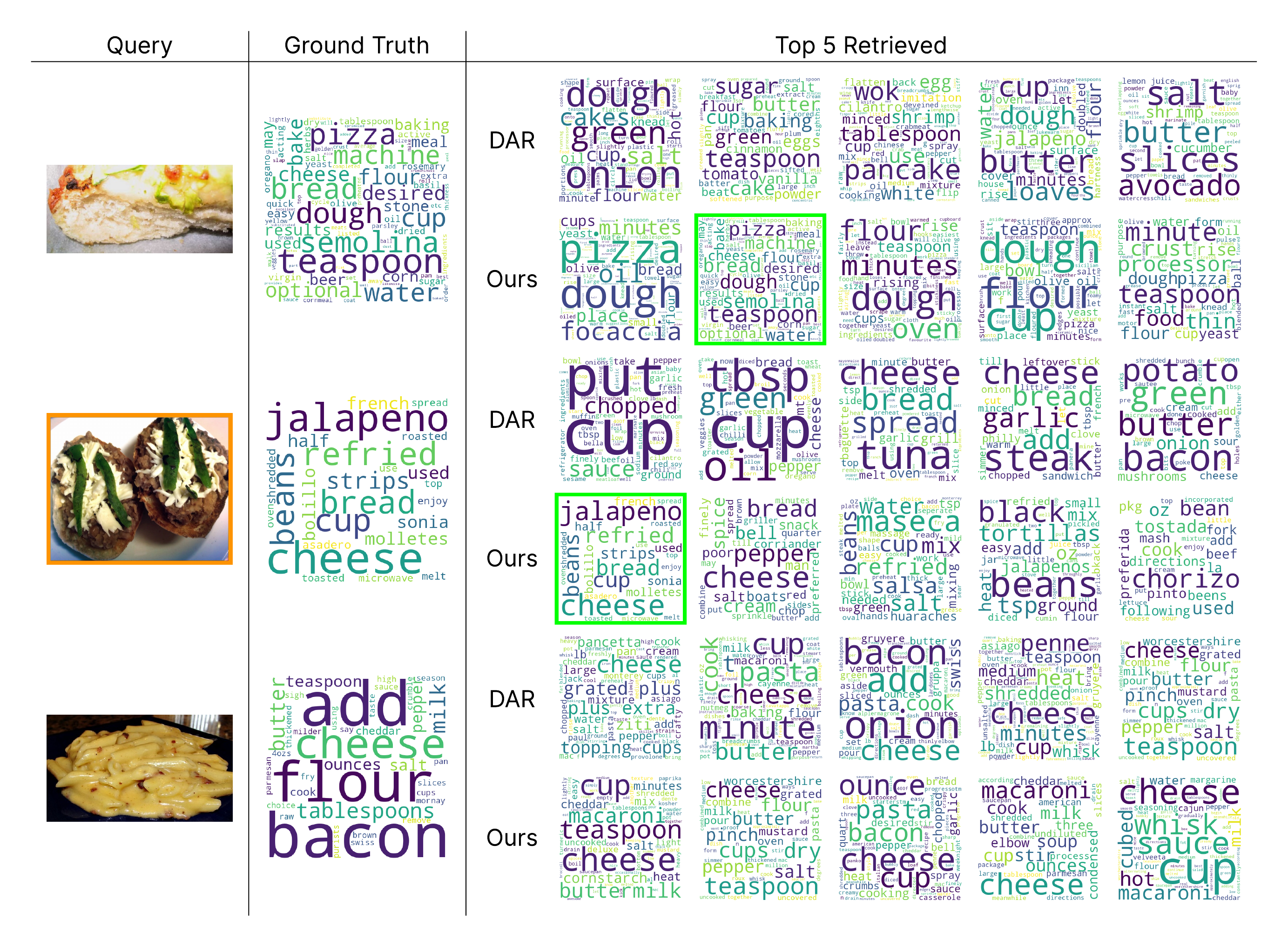}
  \caption{Qualitative comparison of image-to-recipe retrieval between SIMMER (V1-7B) and DAR~\cite{DAR_Song_EnhancingRecipe2024}. For each query image, the top-5 retrieved recipes by both methods are shown, with ground-truth recipes highlighted}
  \label{fig:comparison-i2r}
\end{figure*}

\begin{figure*}
  \centering
  \includegraphics[width=\linewidth]{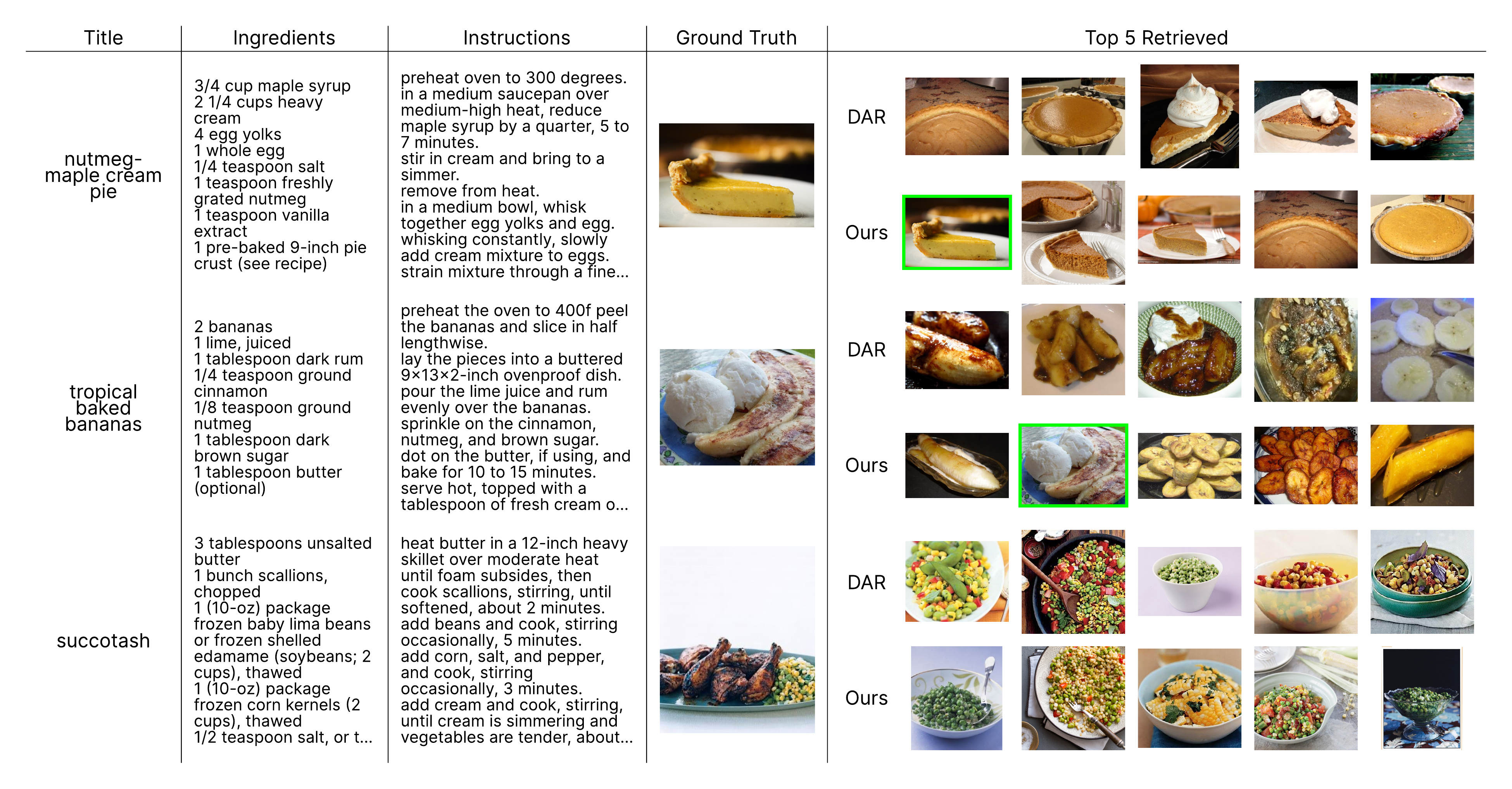}
  \caption{Qualitative comparison of recipe-to-image retrieval between SIMMER (V1-7B) and DAR~\cite{DAR_Song_EnhancingRecipe2024}. For each query recipe, the top-5 retrieved images by both methods are shown, with ground-truth images highlighted}
  \label{fig:comparison-r2i}
\end{figure*}

To further illustrate the advantages of SIMMER, we present a qualitative comparison with DAR~\cite{DAR_Song_EnhancingRecipe2024}, a strong baseline published at ECCV'24 that achieves competitive quantitative performance, as shown in Table~\ref{tab:recipe1m}.
Figures~\ref{fig:comparison-i2r} and~\ref{fig:comparison-r2i} compare the retrieval results of SIMMER (V1-7B) and DAR for the same set of queries on the image-to-recipe and recipe-to-image tasks, respectively.

Figure~\ref{fig:comparison-i2r} presents image-to-recipe retrieval comparisons.
In the top example, SIMMER correctly identifies the toppings on the dough and recognizes the dish as a pizza, retrieving the ground-truth recipe at rank two.
In contrast, DAR misidentifies the dish as a cake, and the correct recipe does not appear within the top five results.
A similar pattern is observed in the middle example: SIMMER recognizes the jalape\~{n}o\footnote{A type of hot chili pepper.} topping on the bread and retrieves the correct recipe at rank one, whereas DAR fails to capture the specific ingredients and does not place the ground-truth recipe in the top five.
The bottom example illustrates a case in which both methods fail.
Because the query image depicts a highly generic macaroni and cheese dish---a category with numerous visually similar instances in the dataset---neither method is able to retrieve the correct recipe within the top five results.

Figure~\ref{fig:comparison-r2i} presents recipe-to-image retrieval comparisons.
In the top example, only SIMMER successfully distinguishes the subtle visual differences between pumpkin pie and the ground-truth nutmeg-maple cream pie, correctly placing the target image at rank one, whereas DAR fails to make this distinction.
In the middle example, SIMMER leverages the textual cue ``slice in half lengthwise'' in the cooking instructions to retrieve the ground-truth image---which depicts a banana sliced lengthwise---at rank two, while DAR does not place it within the top five.
The bottom example represents a failure case caused by a data mismatch that affects both methods: although the query recipe describes succotash\footnote{A traditional American dish primarily consisting of corn and lima beans.}, the annotated ground-truth image prominently features chicken, preventing both methods from retrieving the correct image within the top five results.

\section{Conclusion}

In this paper, we proposed SIMMER, a method for cross-modal retrieval between food images and recipe texts by fine-tuning VLM2Vec, a recent MLLM-based multimodal embedding model, on a recipe dataset.
SIMMER eliminates the need for complex cross-modal alignment objectives and task-specific network architectures, instead leveraging the inherent multimodal understanding capability of a pre-trained MLLM.
We further introduced a prompt design tailored to the structured nature of recipe data for effective processing by the MLLM, as well as a data augmentation strategy that improves robustness to incomplete recipe inputs.
Experiments on the Recipe1M dataset demonstrated that SIMMER achieves state-of-the-art performance, surpassing all prior methods across both the 1k and 10k evaluation settings.

A promising direction for future work is to exploit the ability of the MLLM encoder to jointly process images, recipe texts, and task instructions within a single forward pass.
This opens up possibilities such as retrieving recipes conditioned on both an image and a partial recipe, and enabling instruction-guided retrieval in which the system returns results that reflect user-specified criteria rather than merely the most visually or textually similar dishes.

{
\small
\noindent\textbf{Acknowledgements.}
This work was supported by JSPS KAKENHI 22H00548 and JST CRONOS JPMJCS24K4.
}


{\small
\bibliographystyle{ieeenat_fullname}
\bibliography{simmer}
}

\clearpage
\appendix
\section{Baseline Methods and Full Experimental Results}
\subsection{Baseline Descriptions}
\label{sec:baselines}
We compare SIMMER against the following methods for cross-modal food image--recipe retrieval on the Recipe1M dataset.
All of these methods follow the dual-encoder paradigm, employing separate encoders for food images and recipe texts.

\begin{itemize}
    \item Salvador et al.~\cite{Recipe1M_Salvador_LearningCrossModal2017}: Learns a joint embedding for recipe--image pairs using a margin-based cosine similarity loss over positive/negative pairs, and adds semantic regularization via a shared classifier to align high-level semantics across modalities.
    \item AdaMine~\cite{Adamine_Carvalho_CrossModalRetrieval2018}: Extends joint embedding learning with a double-triplet learning scheme and an adaptive strategy to mine informative triplets, improving convergence and retrieval performance on noisy/variable recipe--image pairs.
    \item Chen et al.~\cite{Chen_DeepUnderstanding2018}: Improves recipe representations by modeling cooking procedures with hierarchical attention (word- and sentence-level) to better align procedure text with visual features for retrieval.
    \item ACME~\cite{ACME_Wang_LearningCrossModal2019}: Proposes ACME, combining an improved triplet-loss training with hard-sample mining, adversarial modality-alignment to match feature distributions, and cross-modal translation consistency via image generation from recipe embeddings and ingredient prediction from image embeddings.
    \item R$^2$GAN~\cite{R2GAN_Zhu_R2GANCrossModal2019}: Introduces R2GAN with one generator and dual discriminators to generate food images from recipe (procedure) text while learning compatible cross-modal features; it also uses a two-level ranking loss in both embedding and image spaces.
    \item DaC~\cite{DaC_Fain_DividingConquering2019}: Proposes a non-parametric retrieval method operating on precomputed image/text embeddings, and shows that extending it with a standard triplet-loss training can reach (or surpass) state-of-the-art on Recipe1M with reduced complexity.
    \item SN~\cite{SN_Zan_SentencebasedNoiserobust2020}: Encodes recipe title, ingredients, and instructions using three attention networks and introduces a sentence-based noise-robust triplet loss to reduce the impact of noisy negative samples in cross-modal alignment.
    \item MCEN~\cite{MCEN_Fu_MCENBridging2020}: Proposes modality-consistent embeddings to bridge the modality gap in image--recipe retrieval, strengthening cross-modal consistency/alignment beyond standard joint embedding losses.
    \item HF-ICMA~\cite{HF-ICMA_Li_HybridFusion2021}: Proposes a hybrid fusion framework that models both intra-modal and cross-modal interactions via attention mechanisms to enhance semantic alignment between food images and structured recipe components.
    \item IMHF~\cite{IMHF_Li_CrossModalImageRecipe2021}: Introduces a lightweight Transformer-based framework that unifies visual and textual feature sequences, enabling both intra- and inter-modality fusion without relying on a separate deep CNN image encoder.
    \item X-MRS~\cite{X-MRS_Guerrero_CrossmodalRetrieval2021}: Uses a transformer-based multilingual recipe encoder and leverages imperfect multilingual translations as a regularizer, improving Recipe1M retrieval and enabling cross-lingual functionality; it also demonstrates conditional image synthesis from recipe embeddings.
    \item M-SIA~\cite{M-SIA_Li_MultisubspaceImplicit2021}: Proposes a multi-subspace implicit alignment framework where both image and recipe embeddings are composed of multiple representation subspaces and are implicitly aligned at the subspace level to exploit fine-grained correspondences.
    \item RDE-GAN~\cite{RDE-GAN_Sugiyama_CrossModalRecipe2021}: Proposes RDE-GAN, disentangling a food image into a recipe-related embedding and a non-recipe “dish shape/style” feature (e.g., serving/plate factors), enabling style-robust retrieval and controllable image generation by recombining embedding and shape/style features.
    \item JEMA~\cite{JEMA_Xie_LearningJoint2021}: Proposes JEMA, a three-tier modality-alignment framework: improve modality-specific encoders, optimize joint embedding with (double) batch-hard triplet loss with soft-margin, and add auxiliary cross-modality regularizers (e.g., category-based alignment and discriminator-based distribution alignment).
    \item SEJE~\cite{SEJE_Xie_LearningTextimage2021}: Proposes SEJE, a two-phase deep feature engineering framework that separates (i) semantic feature engineering in preprocessing (e.g., key-term extraction/scoring using BERT-family models, TextRank, or TF-IDF) from (ii) joint embedding training with optimized batch-hard triplet loss and additional alignment regularizers.
    \item CHEF~\cite{CHEF_Pham_CHEFCrossmodal2021}: Proposes CHEF to model hierarchical relationships among recipe components (title/ingredients/instructions) and their associations to images, using a tree-structured LSTM text encoder to learn more meaningful cross-modal embeddings for retrieval.
    \item H-T~\cite{H-T_Salvador_RevampingCrossModal2021}: Introduces a hierarchical recipe Transformer that encodes recipe components (title/ingredients/instructions) and adds a self-supervised loss over recipe components, enabling training with both paired image--recipe and recipe-only samples for improved retrieval.
    \item SCAN~\cite{SCAN_Wang_CrossModalFood2022}: Proposes SCAN, combining a semantic consistency loss that aligns output semantic probability distributions (e.g., via KL divergence) for paired image/recipe embeddings with a self-attention mechanism to learn more discriminative recipe representations.
    \item EOMA~\cite{EOMA_Xie_CrossModalRetrieval2022}: Proposes an event-oriented modality alignment approach for event-dense text--image retrieval: it models event importance in text (Transformer) and builds image embeddings using event tags/regions to facilitate event-based cross-modal alignment, validated on food/Recipe1M.
    \item Wang et al.~\cite{Wang_PairedCrossModal2022}: Proposes paired cross-modal data augmentation by generating new text--image pairs: it aligns text features with StyleGAN2 latent codes and synthesizes augmented images conditioned on augmented text to boost fine-grained image--text retrieval.
    \item MSJE~\cite{MSJE_Xie_LearningTFIDF2022}: Proposes MSJE, extracting TF-IDF signals from recipe components (title/ingredients/instructions) to enhance multimodal semantics in joint embedding learning for recipe--image retrieval.
    \item Pap et al.~\cite{Papadopoulos_LearningProgram2022}: Represents recipes (and images) as structured cooking programs (graph-like procedural representations) and learns joint embeddings that benefit cross-modal retrieval and related generation/manipulation tasks.
    \item T-Food~\cite{T-Food_Shukor_TransformerDecoders2022}: Proposes T-Food, using Transformer decoders to inject cross-modal interactions as a training-time regularizer while keeping retrieval efficient with unimodal encoders at test time; it also uses dynamic-margin triplet variants and leverages vision--language pre-trained models such as CLIP for the image encoder.
    \item TNLBT~\cite{TNLBT_Yang_TransformerBasedCrossModal2023}: Proposes TNLBT, a Transformer-based cross-modal recipe retrieval framework that leverages large-batch training (and additional modern components such as hierarchical recipe encoding and self-supervised objectives) to improve recipe--image embeddings on Recipe1M.
    \item CIP~\cite{CIP_Huang_ImprovingCrossModal2023}: Adapts CLIP to recipe retrieval through component-aware instance-specific prompt learning, generating prompted visual embeddings without full fine-tuning and integrating structured recipe encoding via a hierarchical Transformer-based merger.
    \item Wang et al.~\cite{Wang_LearningStructural2023}: Learns latent structural (tree) representations for long recipes and incorporates them into a unified framework for recipe generation and food cross-modal retrieval on Recipe1M.
    \item MALM~\cite{MALM_Voutharoja_MALMMask2023}: Proposes MALM, combining token-level local image--text matching with mask augmentation and masked self-distillation to learn more generalizable cross-modal representations for Recipe1M retrieval.
    \item LCWF-GI~\cite{LCWF-GI_Zhao_ExploringLatent2023}: Proposes LCWF-GI, leveraging latent component weight factors and global information to learn more robust recipe and image representations for food-oriented cross-modal retrieval.
    \item MFRF~\cite{MFRF_Li_CrossmodalImageRecipe2023}: Proposes a multimodal fusion framework that encodes structured recipe components with Transformers and BERT-LSTM hybrids, and aligns image--recipe representations using cross-modal attention and contrastive objectives.
    \item CREAMY~\cite{CREAMY_Zou_CREAMYCrossModal2024}: Proposes CREAMY to address matching imperfection in positive image--recipe pairs via Non-Matching and Partial-Matching (NMPM) learning: it avoids forcing full alignment for every positive pair and selectively aligns only matchable parts while exploiting complementary signals from negatives.
    \item Yang et al.~\cite{TNLBT2_Yang_ImprovingCrossModal2024}: Introduces a cross-modal embedding fusing decoder (Cross Decoder) to improve recipe embedding representation, integrates it into a GAN/Transformer-based baseline (TNLBT), and further boosts retrieval with GAN-based image reconstruction and dynamic-margin retrieval losses.
    \item VLPCook~\cite{VLPCook_Shukor_VisionStructuredLanguage2024}: Proposes VLPCook, adapting vision--language pre-training to structured recipe text to inject both local and global recipe contexts into visual representations for improved image--recipe retrieval.
    \item DCA-Food~\cite{DCA-Food_Liu_RevampingImageRecipe2024}: Proposes dual cross-modal attention to strengthen bidirectional interactions between recipe components and image features for improved image--recipe retrieval.
    \item MMACMR~\cite{MMACMR_Zou_DisambiguityAlignment2024}: Proposes MMACMR to handle cross-modal ambiguity by a multi-modal disambiguity and alignment strategy that guides ambiguous image similarity with recipes, while enhancing recipe representations via sentence-level cross-attention between ingredients and instructions.
    \item FARM~\cite{FARM_Wahed_FineGrainedAlignment2024}: Proposes FARM, aligning embeddings of recipe components (title/ingredients/instructions) with image embeddings in a shared space, and introduces a hyperbolic loss (along with triplet training) to better capture class similarity for retrieval.
    \item LMF-CSF~\cite{LMF-CSF_Zhao_EfficientLowrank2024}: Proposes LMF-CSF, an efficient low-rank multi-component fusion method with component-specific factors to fuse recipe components into robust representations for image--recipe retrieval.
    \item UTE-FCL~\cite{UTE-FCL_Zhang_CrossmodalRecipe2024}: Proposes UTE-FCL, using a Unified Text Encoder to efficiently encode the whole recipe (and components) and introducing fine-grained contrastive objectives to better capture component--image correspondences.
    \item DAR~\cite{DAR_Song_EnhancingRecipe2024}: Introduces a data augmentation paradigm using foundation models, where LLM-generated visual imagination descriptions and SAM-based image segments are incorporated into a CLIP-based retrieval framework with adapter layers and multi-level circle loss.
    \item FMI~\cite{FMI_Zhao_CrossModal2025}: Proposes FMI, enhancing recipe representations via cross-component multiscale enriching and improving visual representations via a text-contextualized visual enhancing module embedded into intermediate layers of the image encoder.
    \item Wang et al.~\cite{Wang_MitigatingCrossmodal2025}: Proposes a causal approach that predicts culinary elements potentially overlooked in images and injects them into cross-modal representation learning to mitigate representation bias (evaluated on Recipe1M and multilingual settings).
    \item IR-CLIP~\cite{RecipeRAG_Yang_RecipeRAGAdvancing2025}: RecipeRAG is primarily a recipe generation framework using retrieval-augmented generation and reinforcement fine-tuning; for retrieval, it introduces IR-CLIP by improving CLIP with metric learning and contrastive learning to serve as both retriever and re-ranker for retrieving recipes relevant to a query image.
\end{itemize}

\subsection{Full Comparison with All Baseline Methods}
Tables~\ref{tab:recipe1m-full-1k} and~\ref{tab:recipe1m-full-10k} present the complete comparison of SIMMER with all baseline methods listed in Appendix~\ref{sec:baselines}, including those omitted from the main text for brevity.

\begin{table*}[t]
\caption{Full experimental results on image-to-recipe and recipe-to-image retrieval across all baseline methods (1k setting). The best results are presented in \textbf{bold}, while the next best results are presented in \uline{underlined}}
\label{tab:recipe1m-full-1k}
\begin{adjustbox}{max width=\textwidth}
\begin{tabular}{lccccccccc}
\toprule
\multirow{2}{*}{Method} & \multirow{2}{*}{Venue} & \multicolumn{4}{c}{Image-to-Recipe} & \multicolumn{4}{c}{Recipe-to-Image} \\
\cmidrule(lr){3-6}\cmidrule(lr){7-10}
                        &                        & medR\,$\downarrow$ & R@1\,$\uparrow$ & R@5\,$\uparrow$ & R@10\,$\uparrow$ & medR\,$\downarrow$ & R@1\,$\uparrow$ & R@5\,$\uparrow$ & R@10\,$\uparrow$ \\
\midrule
Salvador et al.~\cite{Recipe1M_Salvador_LearningCrossModal2017} & CVPR'17                 & 5.2  & 24.0          & 51.0          & 65.0          & 5.1  & 25.0          & 52.0          & 65.0          \\
AdaMine~\cite{Adamine_Carvalho_CrossModalRetrieval2018}         & SIGIR'18                & 1.0  & 39.8          & 69.0          & 77.7          & 1.0  & 40.2          & 68.1          & 78.7          \\
Chen et al.~\cite{Chen_DeepUnderstanding2018}                   & MM'18                   & 4.6  & 25.6          & 53.7          & 66.9          & 4.6  & 25.7          & 53.9          & 67.1          \\
ACME~\cite{ACME_Wang_LearningCrossModal2019}                    & CVPR'19                 & 1.0  & 51.8          & 80.2          & 87.5          & 1.0  & 52.8          & 80.2          & 87.6          \\
R$^2$GAN~\cite{R2GAN_Zhu_R2GANCrossModal2019}                   & CVPR'19                 & 2.0  & 39.1          & 71.0          & 81.7          & 2.0  & 40.6          & 72.6          & 83.3          \\
DaC~\cite{DaC_Fain_DividingConquering2019}                      & arXiv'19                   & 1.0  & 60.2          & 84.0          & 89.7          & -    & -             & -             & -             \\
SN~\cite{SN_Zan_SentencebasedNoiserobust2020}                   & ICMR'20                 & 1.0  & 52.7          & 81.7          & 88.9          & 1.0  & 54.1          & 81.8          & 88.9          \\
MCEN~\cite{MCEN_Fu_MCENBridging2020}                            & CVPR'20                 & 2.0  & 48.2          & 75.8          & 83.6          & 1.9  & 48.4          & 76.1          & 83.7          \\
HF-ICMA~\cite{HF-ICMA_Li_HybridFusion2021}                      & SIGIR'21                & 1.0  & 55.1          & 86.7          & 92.4          & 1.0  & 56.8          & 87.5          & 93.0          \\
IMHF~\cite{IMHF_Li_CrossModalImageRecipe2021}                   & ICMR'21                 & 1.0  & 53.2          & 80.7          & 87.6          & 1.0  & 54.1          & 82.4          & 88.2          \\
X-MRS~\cite{X-MRS_Guerrero_CrossmodalRetrieval2021}             & MM'21                   & 1.0  & 64.0          & 88.3          & 92.6          & 1.0  & 63.9          & 87.6          & 92.6          \\
M-SIA~\cite{M-SIA_Li_MultisubspaceImplicit2021}                 & CIKM'21                 & 1.0  & 59.3          & 86.3          & 92.6          & 1.0  & 59.8          & 86.7          & 92.8          \\
RDE-GAN~\cite{RDE-GAN_Sugiyama_CrossModalRecipe2021}            & MM'21                   & 1.0  & 59.4          & 81.0          & 87.4          & 1.0  & 61.2          & 81.0          & 87.2          \\
JEMA~\cite{JEMA_Xie_LearningJoint2021}                          & CIKM'21                 & 1.0  & 58.1          & 85.8          & 92.2          & 1.0  & 58.5          & 86.2          & 92.3          \\
SEJE~\cite{SEJE_Xie_LearningTextimage2021}                      & ACM Trans. Inf. Syst.'21& 1.0  & 58.1          & 85.8          & 92.2          & 1.0  & 58.5          & 86.2          & 92.3          \\
CHEF~\cite{CHEF_Pham_CHEFCrossmodal2021}                        & AAAI'21                 & 1.8  & 49.4          & 79.6          & 86.1          & 1.5  & 49.8          & 79.0          & 86.4          \\
H-T~\cite{H-T_Salvador_RevampingCrossModal2021}                 & CVPR'21                 & 1.0  & 60.0          & 87.6          & 92.9          & 1.0  & 60.3          & 87.6          & 93.2          \\
SCAN~\cite{SCAN_Wang_CrossModalFood2022}                        & IEEE TMM'22            & 1.0  & 54.0          & 81.9          & 89.2          & 1.0  & 54.9          & 81.9          & 89.0          \\
EOMA~\cite{EOMA_Xie_CrossModalRetrieval2022}                    & ICMR'22                 & 1.0  & 77.5          & 94.1          & 96.8          & 1.0  & 77.5          & 94.0          & 96.7          \\
Wang et al.~\cite{Wang_PairedCrossModal2022}                    & MM'22                   & -    & 60.6          & 87.7          & 92.8          & -    & 61.3          & 87.7          & 93.2          \\
MSJE~\cite{MSJE_Xie_LearningTFIDF2022}                          & IEEE TSC'22            & 1.0  & 56.5          & 84.7          & 90.9          & 1.0  & 56.2          & 84.9          & 91.1          \\
Pap et al.~\cite{Papadopoulos_LearningProgram2022}              & CVPR'22                 & 1.0  & 66.9          & 90.9          & 95.1          & 1.0  & 66.8          & 89.8          & 94.6          \\
T-Food~\cite{T-Food_Shukor_TransformerDecoders2022}             & CVPRW'22                & 1.0  & 72.3          & 90.7          & 93.4          & 1.0  & 72.6          & 90.6          & 93.4          \\
TNLBT~\cite{TNLBT_Yang_TransformerBasedCrossModal2023}          & MMM'23                  & 1.0  & 81.0          & 95.2          & 97.4          & 1.0  & 80.3          & 95.2          & 97.4          \\
CIP~\cite{CIP_Huang_ImprovingCrossModal2023}                    & MM'23                   & 1.0  & 77.1          & 94.2          & 97.2          & 1.0  & 77.3          & 94.4          & 97.0          \\
Wang et al.~\cite{Wang_LearningStructural2023}                  & IEEE TPAMI'23          & 1.0  & 53.5          & 81.5          & 88.8          & 1.0  & 55.0          & 82.0          & 88.8          \\
MALM~\cite{MALM_Voutharoja_MALMMask2023}                        & arXiv'23                   & 1.0  & 74.0          & 91.3          & 94.3          & 1.0  & 73.0          & 91.0          & 93.9          \\
LCWF-GI~\cite{LCWF-GI_Zhao_ExploringLatent2023}                 & Connection Science'23  & 1.0  & 59.4          & 86.8          & 92.5          & 1.0  & 60.1          & 86.7          & 92.7          \\
MFRF~\cite{MFRF_Li_CrossmodalImageRecipe2023}                   & MMAsia'23               & 1.0  & 62.3          & 86.9          & 91.5          & 1.0  & 64.8          & 88.7          & 92.7          \\
CREAMY~\cite{CREAMY_Zou_CREAMYCrossModal2024}                   & IEEE Access'24         & 1.0  & 73.3          & 92.5          & 95.6          & 1.0  & 73.2          & 92.5          & 95.8          \\
Yang et al.~\cite{TNLBT2_Yang_ImprovingCrossModal2024}          & ICMRW'24                & 1.0  & \uline{81.8}  & \uline{95.9}  & \uline{97.8}  & 1.0  & \uline{81.2}  & \uline{96.0}  & \uline{97.9}  \\
VLPCook~\cite{VLPCook_Shukor_VisionStructuredLanguage2024}      & CVIU'24                & 1.0  & 73.6          & 90.5          & 93.3          & 1.0  & 74.7          & 90.7          & 93.2          \\
DCA-Food~\cite{DCA-Food_Liu_RevampingImageRecipe2024}           & Mathematics'24         & 1.0  & 70.9          & 89.6          & 93.5          & 1.0  & 70.8          & 90.1          & 93.9          \\
MMACMR~\cite{MMACMR_Zou_DisambiguityAlignment2024}              & Foods'24               & 1.0  & 69.1          & 90.8          & 94.9          & 1.0  & 69.2          & 90.6          & 95.0          \\
FARM~\cite{FARM_Wahed_FineGrainedAlignment2024}                 & WACV'24                 & 1.0  & 73.7          & 90.7          & 93.4          & 1.0  & 73.6          & 90.8          & 93.5          \\
LMF-CSF~\cite{LMF-CSF_Zhao_EfficientLowrank2024}                & Multimedia Tools Appl.'24& 1.0  & 65.8          & 89.7          & 94.3          & 1.0  & 65.5          & 89.4          & 94.3          \\
UTE-FCL~\cite{UTE-FCL_Zhang_CrossmodalRecipe2024}               & Knowledge-Based Systems'24& 1.0  & 68.1          & 90.0          & 94.5          & 1.0  & 67.2          & 90.1          & 94.7          \\
DAR~\cite{DAR_Song_EnhancingRecipe2024}                         & ECCV'24                 & 1.0  & 77.3          & 95.3          & 97.7          & 1.0  & 77.1          & 95.4          & \uline{97.9}  \\
FMI~\cite{FMI_Zhao_CrossModal2025}                              & Scientific Reports'25  & 1.0  & 77.4          & 95.8          & 97.6          & 1.0  & 77.1          & 95.4          & 97.7          \\
Wang et al.~\cite{Wang_MitigatingCrossmodal2025}                & MM'25                   & 1.0  & 79.1          & 94.6          & 97.0          & 1.0  & 78.3          & 95.0          & 97.2          \\
IR-CLIP~\cite{RecipeRAG_Yang_RecipeRAGAdvancing2025}            & MM'25                   & -    & -             & -             & -             & -    & -             & -             & -             \\
\midrule
SIMMER (V1-2B)                                                    & -                       & 1.0  & 84.1          & 97.3          & 98.8          & 1.0  & 81.5          & 96.6          & 98.4          \\
SIMMER (V1-7B)                                                    & -                       & 1.0  & \textbf{87.5} & \textbf{98.0} & \textbf{99.2} & 1.0  & \textbf{85.1} & \textbf{97.6} & \textbf{99.1} \\
SIMMER (V2)                                                       & -                       & 1.0  & 83.8          & 96.9          & 98.7          & 1.0  & 81.7          & 96.5          & 98.3          \\
\bottomrule
\end{tabular}
\end{adjustbox}
\end{table*}

\begin{table*}[t]
\caption{Full experimental results on image-to-recipe and recipe-to-image retrieval across all baseline methods (10k setting). The best results are presented in \textbf{bold}, while the next best results are presented in \uline{underlined}}
\label{tab:recipe1m-full-10k}
\begin{adjustbox}{max width=\textwidth}
\begin{tabular}{lccccccccc}
\toprule
\multirow{2}{*}{Method} & \multirow{2}{*}{Venue} & \multicolumn{4}{c}{Image-to-Recipe} & \multicolumn{4}{c}{Recipe-to-Image} \\
\cmidrule(lr){3-6}\cmidrule(lr){7-10}
                        &                        & medR\,$\downarrow$ & R@1\,$\uparrow$ & R@5\,$\uparrow$ & R@10\,$\uparrow$ & medR\,$\downarrow$ & R@1\,$\uparrow$ & R@5\,$\uparrow$ & R@10\,$\uparrow$ \\
\midrule
Salvador et al.~\cite{Recipe1M_Salvador_LearningCrossModal2017} & CVPR'17                 & 41.9 & -             & -             & -             & 39.2 & -             & -             & -             \\
AdaMine~\cite{Adamine_Carvalho_CrossModalRetrieval2018}         & SIGIR'18                & 13.2 & 14.9          & 35.3          & 45.2          & 12.2 & 14.8          & 34.6          & 46.1          \\
Chen et al.~\cite{Chen_DeepUnderstanding2018}                   & MM'18                   & 39.8 & 7.2           & 19.2          & 27.6          & 38.1 & 7.0           & 19.4          & 27.8          \\
ACME~\cite{ACME_Wang_LearningCrossModal2019}                    & CVPR'19                 & 6.7  & 22.9          & 46.8          & 57.9          & 6.0  & 24.4          & 47.9          & 59.0          \\
R$^2$GAN~\cite{R2GAN_Zhu_R2GANCrossModal2019}                   & CVPR'19                 & 13.9 & 13.5          & 33.5          & 44.9          & 12.6 & 14.2          & 35.0          & 46.8          \\
DaC~\cite{DaC_Fain_DividingConquering2019}                      & arXiv'19                   & 4.0  & 30.0          & 56.5          & 67.0          & -    & -             & -             & -             \\
SN~\cite{SN_Zan_SentencebasedNoiserobust2020}                   & ICMR'20                 & 7.0  & 22.1          & 45.9          & 56.9          & 7.0  & 23.4          & 47.3          & 57.9          \\
MCEN~\cite{MCEN_Fu_MCENBridging2020}                            & CVPR'20                 & 7.2  & 20.3          & 43.3          & 54.4          & 6.6  & 21.4          & 44.3          & 55.2          \\
HF-ICMA~\cite{HF-ICMA_Li_HybridFusion2021}                      & SIGIR'21                & 5.0  & 24.0          & 51.6          & 65.4          & 4.2  & 25.6          & 54.8          & 67.3          \\
IMHF~\cite{IMHF_Li_CrossModalImageRecipe2021}                   & ICMR'21                 & 6.2  & 23.4          & 48.2          & 58.4          & 5.8  & 24.9          & 48.3          & 59.4          \\
X-MRS~\cite{X-MRS_Guerrero_CrossmodalRetrieval2021}             & MM'21                   & 3.0  & 32.9          & 60.6          & 71.2          & 3.0  & 33.0          & 60.4          & 70.7          \\
M-SIA~\cite{M-SIA_Li_MultisubspaceImplicit2021}                 & CIKM'21                 & 4.0  & 29.2          & 55.0          & 66.2          & 4.0  & 30.3          & 55.6          & 66.5          \\
RDE-GAN~\cite{RDE-GAN_Sugiyama_CrossModalRecipe2021}            & MM'21                   & 3.5  & 36.0          & 56.1          & 64.4          & 3.0  & 38.2          & 57.7          & 65.8          \\
JEMA~\cite{JEMA_Xie_LearningJoint2021}                          & CIKM'21                 & 4.2  & 26.9          & 54.0          & 65.6          & 4.0  & 27.2          & 54.4          & 66.1          \\
SEJE~\cite{SEJE_Xie_LearningTextimage2021}                      & ACM Trans. Inf. Syst.'21& 4.2  & 26.9          & 54.0          & 65.6          & 4.0  & 27.2          & 54.4          & 66.1          \\
CHEF~\cite{CHEF_Pham_CHEFCrossmodal2021}                        & AAAI'21                 & 7.3  & 20.9          & 44.8          & 56.3          & 7.0  & 21.9          & 45.3          & 56.6          \\
H-T~\cite{H-T_Salvador_RevampingCrossModal2021}                 & CVPR'21                 & 4.0  & 27.9          & 56.4          & 68.1          & 4.0  & 28.3          & 56.5          & 68.1          \\
SCAN~\cite{SCAN_Wang_CrossModalFood2022}                        & IEEE TMM'22            & 5.9  & 23.7          & 49.3          & 60.6          & 5.1  & 25.3          & 50.6          & 61.6          \\
EOMA~\cite{EOMA_Xie_CrossModalRetrieval2022}                    & ICMR'22                 & 1.0  & 50.6          & 77.1          & 84.8          & 1.3  & 50.1          & 76.8          & 84.5          \\
Wang et al.~\cite{Wang_PairedCrossModal2022}                    & MM'22                   & -    & 28.6          & 57.1          & 68.6          & -    & 29.3          & 57.3          & 69.0          \\
MSJE~\cite{MSJE_Xie_LearningTFIDF2022}                          & IEEE TSC'22            & 5.0  & 25.6          & 52.1          & 63.8          & 5.0  & 26.2          & 52.5          & 64.1          \\
Pap et al.~\cite{Papadopoulos_LearningProgram2022}              & CVPR'22                 & -    & -             & -             & -             & -    & -             & -             & -             \\
T-Food~\cite{T-Food_Shukor_TransformerDecoders2022}             & CVPRW'22                & 2.0  & 43.4          & 70.7          & 79.7          & 2.0  & 44.6          & 71.2          & 79.7          \\
TNLBT~\cite{TNLBT_Yang_TransformerBasedCrossModal2023}          & MMM'23                  & 1.0  & \uline{56.5}  & 80.7          & 87.1          & 1.0  & \uline{55.9}  & 80.1          & 86.8          \\
CIP~\cite{CIP_Huang_ImprovingCrossModal2023}                    & MM'23                   & 2.0  & 44.9          & 72.8          & 82.0          & 2.0  & 45.2          & 73.0          & 81.8          \\
Wang et al.~\cite{Wang_LearningStructural2023}                  & IEEE TPAMI'23          & 6.0  & 23.4          & 48.8          & 60.1          & 5.6  & 24.6          & 50.0          & 61.0          \\
MALM~\cite{MALM_Voutharoja_MALMMask2023}                        & arXiv'23                   & 2.0  & 45.9          & 72.3          & 80.5          & 2.0  & 44.2          & 71.7          & 80.1          \\
LCWF-GI~\cite{LCWF-GI_Zhao_ExploringLatent2023}                 & Connection Science'23  & 4.0  & 27.9          & 56.0          & 67.8          & 4.0  & 28.6          & 55.8          & 67.5          \\
MFRF~\cite{MFRF_Li_CrossmodalImageRecipe2023}                   & MMAsia'23               & 5.0  & 31.2          & 56.5          & 67.3          & 4.4  & 33.7          & 59.3          & 68.5          \\
CREAMY~\cite{CREAMY_Zou_CREAMYCrossModal2024}                   & IEEE Access'24         & 2.0  & 44.6          & 71.6          & 80.4          & 2.0  & 45.0          & 71.4          & 80.0          \\
Yang et al.~\cite{TNLBT2_Yang_ImprovingCrossModal2024}          & ICMRW'24                & 1.0  & \uline{56.5}  & \uline{81.0}  & \uline{87.6}  & 1.0  & 55.7          & \uline{80.2}  & \uline{87.1}  \\
VLPCook~\cite{VLPCook_Shukor_VisionStructuredLanguage2024}      & CVIU'24                & 2.0  & 45.3          & 72.4          & 80.8          & 2.0  & 46.4          & 73.1          & 80.9          \\
DCA-Food~\cite{DCA-Food_Liu_RevampingImageRecipe2024}           & Mathematics'24         & 2.0  & 42.7          & 68.9          & 77.6          & 2.0  & 42.9          & 69.7          & 78.5          \\
MMACMR~\cite{MMACMR_Zou_DisambiguityAlignment2024}              & Foods'24               & 2.1  & 38.1          & 65.8          & 75.9          & 2.2  & 38.3          & 65.6          & 75.6          \\
FARM~\cite{FARM_Wahed_FineGrainedAlignment2024}                 & WACV'24                 & 2.0  & 44.9          & 71.8          & 80.0          & 2.0  & 44.3          & 71.5          & 80.0          \\
LMF-CSF~\cite{LMF-CSF_Zhao_EfficientLowrank2024}                & Multimedia Tools Appl.'24& 3.0  & 34.6          & 62.7          & 73.2          & 3.0  & 34.3          & 62.5          & 72.8          \\
UTE-FCL~\cite{UTE-FCL_Zhang_CrossmodalRecipe2024}               & Knowledge-Based Systems'24& 2.7  & 37.4          & 65.4          & 75.4          & 3.0  & 36.5          & 64.7          & 74.8          \\
DAR~\cite{DAR_Song_EnhancingRecipe2024}                         & ECCV'24                 & 2.0  & 47.8          & 75.9          & 84.3          & 2.0  & 47.4          & 75.5          & 84.1          \\
FMI~\cite{FMI_Zhao_CrossModal2025}                              & Scientific Reports'25  & 1.0  & 48.4          & 76.3          & 81.9          & 1.0  & 49.5          & 79.2          & 83.1          \\
Wang et al.~\cite{Wang_MitigatingCrossmodal2025}                & MM'25                   & 1.0  & 51.7          & 78.2          & 85.9          & 1.0  & 52.2          & 78.4          & 86.0          \\
IR-CLIP~\cite{RecipeRAG_Yang_RecipeRAGAdvancing2025}            & MM'25                   & 2.0  & 46.5          & 74.3          & 83.1          & -    & -             & -             & -             \\
\midrule
SIMMER (V1-2B)                                                    & -                       & 1.0  & 59.7          & 83.5          & 89.9          & 1.0  & 55.8          & 81.1          & 88.0          \\
SIMMER (V1-7B)                                                    & -                       & 1.0  & \textbf{65.5} & \textbf{87.4} & \textbf{92.5} & 1.0  & \textbf{61.5} & \textbf{85.0} & \textbf{91.0} \\
SIMMER (V2)                                                       & -                       & 1.0  & 59.1          & 83.3          & 89.8          & 1.0  & 55.7          & 81.0          & 88.0          \\
\bottomrule
\end{tabular}
\end{adjustbox}
\end{table*}

\end{document}